%% file: arxiv.tex
\documentclass{article} 
\PassOptionsToPackage{table}{xcolor}
\usepackage{iclr2027_conference,times}
\input{math_commands.tex}

\usepackage{url}
\usepackage[table]{xcolor}
\usepackage{microtype}
\usepackage{graphicx}
\usepackage{subfigure}
\usepackage{booktabs} 
\usepackage{multirow}
\usepackage{makecell}
\usepackage{algorithm}
\usepackage{algorithmic}
\usepackage{wrapfig}
\usepackage{bm}
\usepackage{arydshln}
\usepackage{svg}
\usepackage{makecell}
\usepackage{pifont}

\usepackage{amsmath}
\usepackage{amssymb}
\usepackage{mathtools}
\usepackage{amsthm}
\usepackage{enumitem}

\theoremstyle{plain}

\theoremstyle{definition}

\theoremstyle{remark}

\usepackage[textsize=tiny]{todonotes}

\usepackage[utf8]{inputenc} 
\usepackage[T1]{fontenc}    
\usepackage{amsfonts}       
\usepackage{nicefrac}       
\usepackage{hyperref}
\usepackage{colortbl}
\newsavebox{\dynscmetricbox}

\usepackage[most]{tcolorbox}
\tcbuselibrary{breakable,skins}
\definecolor{ConsBenchPromptBlue}{HTML}{EDF5FF}
\definecolor{ConsBenchPromptBorder}{HTML}{89AFD8}
\definecolor{ConsBenchPromptTitle}{HTML}{204E7A}
\newtcolorbox{consbenchprompt}[1]{
  enhanced,
  breakable,
  colback=ConsBenchPromptBlue,
  colframe=ConsBenchPromptBorder,
  colbacktitle=ConsBenchPromptTitle,
  coltitle=white,
  title={Prompt: #1},
  fonttitle=\small\bfseries,
  fontupper=\small,
  boxrule=0.5pt,
  arc=2mm,
  left=3mm,right=3mm,top=2.5mm,bottom=2.5mm,
  toptitle=1.5mm,bottomtitle=1.5mm,
  before skip=12pt,after skip=12pt,
  parbox=false,
  before upper={\setlength{\parindent}{0pt}\raggedright},
  title after break={Prompt: #1 (continued)}
}
\title{Towards Subject Consistency over Dynamic Subject Sets in Video Generation}

\author{Tongcheng Zhang,\quad Jun Zhu,\quad Jianfei Chen\thanks{Corresponding author.} \\
\normalfont Dept. of Comp. Sci. and Tech., Institute for AI, BNRist Center, THBI Lab, \\
\normalfont Tsinghua-Bosch Joint ML Center, Tsinghua University \\
\normalfont\small\texttt{zhangtc26@mails.tsinghua.edu.cn, \{dcszj, jianfeic\}@tsinghua.edu.cn}
}

\iclrfinalcopy
\makeatletter
\let\arxiv@maketitle\@maketitle
\renewcommand{\@maketitle}{\arxiv@maketitle\fancyhead{}\lhead{Preprint.}}
\makeatother
\renewcommand{\headrulewidth}{0.4pt}
\hypersetup{
  pdftitle={Towards Subject Consistency over Dynamic Subject Sets in Video Generation},
  pdfauthor={Tongcheng Zhang, Jun Zhu, Jianfei Chen}
}
\begin{document}
\maketitle

\begin{abstract}
We argue that as video generation extends to longer durations, subject consistency should be evaluated over \textit{dynamic subject sets}. We therefore introduce \textbf{DynSC-Eval}, an evaluation framework that dynamically tracks eligible subjects throughout their visible lifespans and measures local continuity and global identity preservation using six complementary object-level metrics, with explicit detection of inconsistency events. To validate its effectiveness, we design synthetic experiments that actively inject inconsistency events, demonstrating both the sensitivity of DynSC-Eval and the limitations of existing metrics. Evaluations of diverse models on 5s, 15s, and 60s video generation further reveal substantial subject consistency differences that are obscured by conventional metrics. Beyond evaluation, we construct rewards from DynSC-Eval and apply DiffusionNFT post-training in an autonomous-driving testbed. On 5s generation, our approach reduces the six inconsistency metrics by an average of 13.82\% for Wan-2.1-1.3B and 5.66\% for SANA-2B, with improvements also observed on the I2V model ReSim. Qualitative comparisons further demonstrate the effectiveness of our method.
 We then extend generation to 10s and 30s through curriculum learning and show that consistency optimization remains effective while largely preserving other capabilities.
\end{abstract}

\begin{figure}[h]
    \centering
    \includegraphics[width=\textwidth]{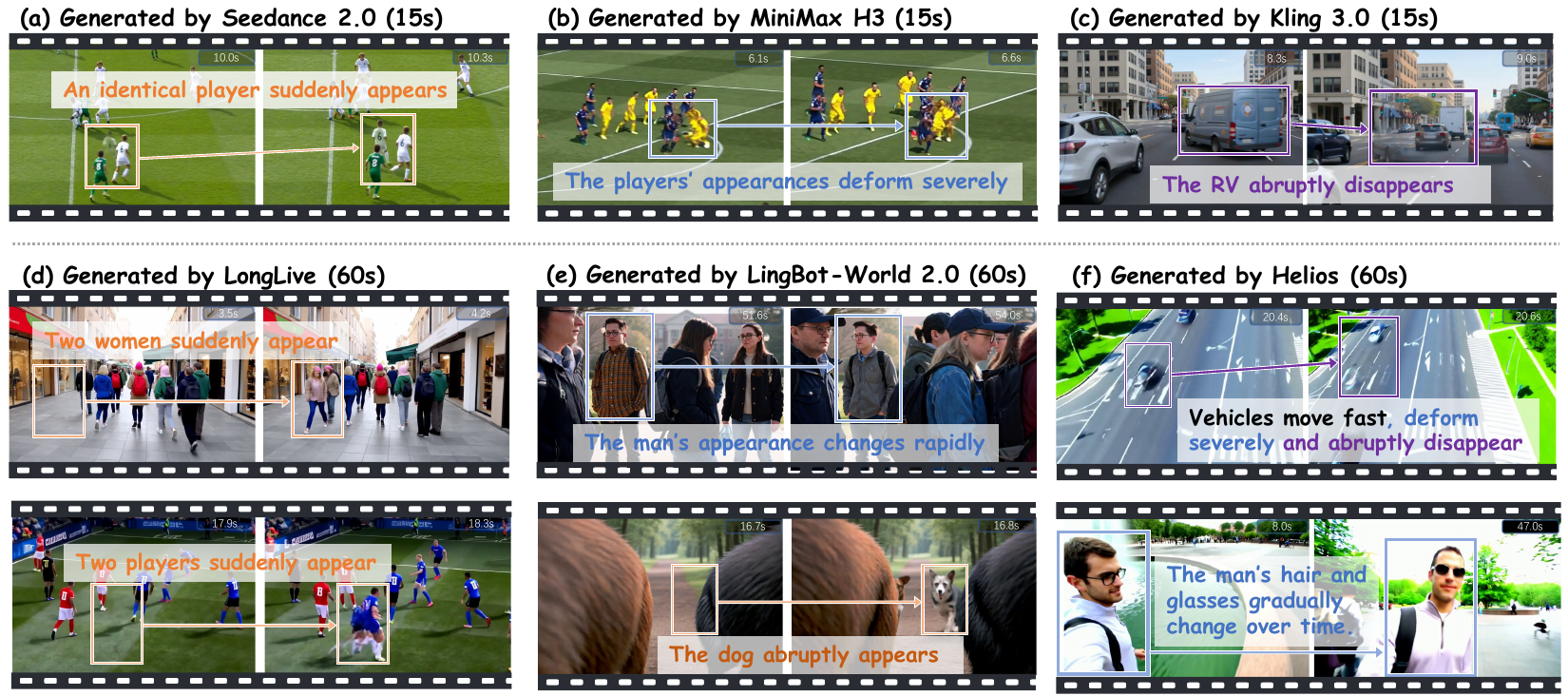}
    \caption{\textbf{Examples of subject inconsistency in long video generation.} As the generation duration increases, subject inconsistency events become increasingly prevalent. The  corresponding prompts are provided in Appendix~\ref{app:video-generation-prompts}.  Additional results are presented on the \href{https://dynsc-paper.pages.dev/}{project website}.}
    \label{fig:failure-mode}
\end{figure}

\section{Introduction}
Video models are increasingly moving towards long-video generation. Many commercial models now natively support generation of high-quality videos more than 10 seconds~\citep{openai2025sora2,Wiedemer2025VideoMA,Team2025KlingOmniTR,Seedance2026Seedance2A,minimax2026h3}, while some streaming models can even achieve minute-level video generation~\citep{Chen2025SkyReelsV2IF, Yang2025LongLiveRI, Yuan2026HeliosRR, Gao2026InfiniteWW, Zhu2026SANAWMEM}.
Over longer temporal horizons, subjects may enter, leave, and interact, allowing both the composition of the visible subject set and the relationships among its members to evolve.
Even within a continuous single shot, maintaining temporal consistency across these changing subjects remains challenging.
Fig.~\ref{fig:failure-mode} demonstrates failures in long-video generation.

The frequency and severity of these failures can reveal differences in models' capability. An ideal video generator should maintain each subject's temporal consistency throughout its visible lifespan. Thus, \textit{evaluating subject consistency over dynamic subject sets} is a more appropriate assessment way.
 Here, \textit{dynamic subject sets} means both the membership and cardinality of the visible subject set can change over time. Evaluation of subject consistency should cover the union of dynamic subject sets throughout the video. However, existing consistency metrics primarily rely on holistic frame-level comparisons~\citep{Huang2023VBenchCB,Liu2023EvalCrafterBA, Huang2024VBenchCA,Liu2026LongAVCompassTU}, sparsely sampled frames~\citep{Sun2024T2VCompBenchAC}, or reference-based assessment of selected subjects~\citep{Zhou2026DrivingGenAC, Zhang2026MBenchAC, Ying2026WBenchAC, li2026joyai}. These approaches either lack sensitivity in subject detection or overlook inconsistencies involving subjects that enter later in the video.

To address this gap, we introduce \textbf{DynSC-Eval}, an evaluation framework that dynamically updates the subject set to track each eligible subject throughout its visible lifespan. It decomposes temporal consistency into local and global components, measured using six complementary metrics. Local consistency captures short-term temporal continuity and incorporates explicit detection and counting of inconsistency events, while global consistency measures long-term identity preservation. 

We design a controlled synthetic experiment that intentionally injects local and global inconsistencies. In this setting, evaluating only subjects visible in the initial frames cannot fully capture subject consistency throughout the video. Meanwhile, compared with holistic frame-level evaluation like VBench, our metrics show substantially greater sensitivity and clearly reflect changes in the prevalence of injected inconsistencies.
We further validate DynSC-Eval across a diverse set of commercial and open-source models at 5s, commercial models at 15s, and streaming models at 60s.
 DynSC-Eval distinguishes models with similar VBench consistency scores and reveals complementary failure patterns in local continuity and global identity preservation, demonstrating its effectiveness.

\begin{figure}[tb]
    \centering
    \includegraphics[width=\textwidth]{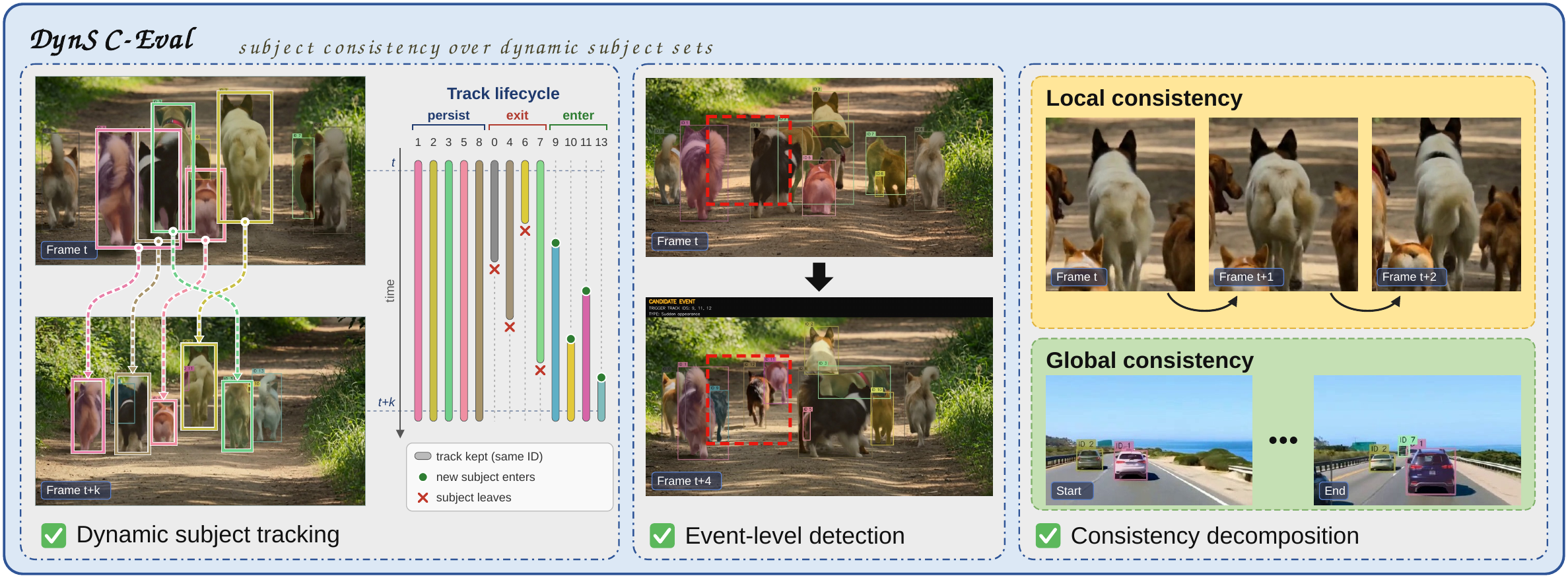}
    \caption{\textbf{Overview of DynSC-Eval.}  The framework dynamically tracks subjects and their lifecycle, detects inconsistency events (e.g., sudden appearance/disappearance), and decomposes consistency into local and global consistency, enabling fine-grained evaluation of dynamic subject sets.}
    \label{fig:VConsbench}
\end{figure}

Furthermore, we design rewards based on DynSC-Eval to guide model optimization and improve subject consistency. We adopt autonomous driving as a representative testbed, where high-speed motion with multiple objects tend to expose temporal inconsistencies, which is practically critical for this domain. We first fine-tune text-to-video (T2V) models Wan-2.1-1.3B~\citep{Wang2025WanOA} and SANA-2B~\citep{Chen2025SANAVideoEV} on 100 hours of real-world videos to establish 5s baselines. We then design consistency rewards based on DynSC-Eval metrics and perform post-training with DiffusionNFT~\citep{Zheng2025DiffusionNFTOD}. On an independent held-out evaluation set of 100 videos, our approach yields average relative reductions of $\textbf{13.82\%}$ and $\textbf{5.66\%}$ across all six inconsistency metrics for Wan-2.1-1.3B and SANA-2B, respectively, while other capabilities, such as visual quality and motion dynamics, are preserved. Qualitative results 
also align with these numerical findings. We further apply the same optimization framework to the image-to-video (I2V) model ReSim~\citep{Yang2025ReSimRW} and improve all consistency metrics of DynSC-Eval. These results demonstrate the effectiveness of our consistency rewards across both T2V and I2V generation.

Building on the improvements achieved for 5s generation, we extend our study to longer temporal horizons. Through curriculum learning, we progressively adapt the model to 10s and then 30s generation using an additional 100 and 200 hours of OpenDV data, respectively. We then apply DiffusionNFT with our consistency reward to both models, improving DynSC at both durations while largely preserving other capabilities.  Our contributions are summarized as follows:

(i) We highlight the importance that subject consistency evaluation should account for dynamic subject sets as video generation extends to longer durations.

(ii) We introduce \textbf{DynSC-Eval}, an evaluation framework  that combines dynamic subject tracking, event-level detection, and consistency decomposition. We validate its sensitivity through synthetic experiments and demonstrate its effectiveness across diverse models and video durations.

(iii) We design rewards based on DynSC-Eval and adopt autonomous driving as a testbed to post-train 5s models, achieving substantial subject consistency gains across both T2V and I2V models while largely preserving other capabilities, with qualitative results further supporting these improvements.

(iv) We then extend video generation from 5s to 30s via curriculum learning, demonstrating that our reward optimization also yields significant consistency gains in long-video generation.

\section{Related work}
\label{a_sec:related works}
\subsection{Temporal Consistency Evaluation}

Temporal consistency is an important dimension of video generation evaluation. General benchmarks, including VBench~\citep{Huang2023VBenchCB}, VBench++~\citep{Huang2024VBenchCA}, and EvalCrafter~\citep{Liu2023EvalCrafterBA}, primarily assess temporal consistency through holistic frame-level comparisons. A complementary line uses human judgments and learned evaluators~\citep{ He2024VideoScoreBA, Han2025VideoBenchHV, Li2025WorldModelBenchJV, Inbasekar2026WorldJenAE}.
LongAV-Compass~\citep{Liu2026LongAVCompassTU} extends evaluation to minute-level audio-visual generation using MLLM assessment and complementary perceptual metrics. These approaches provide broad perceptual and semantic coverage. However, such holistic evaluations struggle to localize inconsistencies in individual subjects and have limited sensitivity to discrete inconsistency events.

Subject-level evaluations address more specific aspects of consistency. T2V-CompBench~\citep{Sun2024T2VCompBenchAC} evaluates compositional generation via MLLM judges, subject detection, and tracking, targeting consistent attribute binding and object interactions. JoyAI-Echo~\citep{li2026joyai} emphasizes character identity consistency across shots in long-form audio-visual generation. Recent works, including DrivingGen~\citep{Zhou2026DrivingGenAC}, MBench~\citep{Zhang2026MBenchAC}, and WBench~\citep{Ying2026WBenchAC}, evaluate subject consistency using designated targets or first-frame-based protocols, yet comprehensive coverage of the evolving subject set, including subjects entering later as videos grow longer, remains insufficiently addressed.
In contrast, DynSC-Eval explicitly evaluates dynamic subject sets, including all eligible subjects that first appear after the initial frame. It separately measures local disruptions and global drift and detects inconsistency events, providing a more comprehensive and accurate assessment of subject consistency throughout the video.

\subsection{Post-Training for Temporal Consistency}
Policy optimization methods such as PPO~\citep{Schulman2017ProximalPO} and GRPO~\citep{Shao2024DeepSeekMathPT} provide foundations for learning from reward feedback. For diffusion models, DiffusionNFT~\citep{Zheng2025DiffusionNFTOD} incorporates reward feedback into forward-process flow-matching updates by contrasting positive and negative generations. In video generation, InstructVideo~\citep{Yuan2023InstructVideoIV} adapts image-based preference rewards through segmental sampling and temporal attenuation while Control-A-Video~\citep{Chen2023ControlAVideoCT} incorporates reward feedback into controllable generation. Some works explicitly design consistency signals to improve temporal consistency in video generation. Video Consistency Distance~\citep{Aoshima2025VideoCD} uses a frequency-domain discrepancy between generated frame features and a conditioning image for I2V fine-tuning. IPRO~\citep{Shen2025IdentityPreservingIG} targets facial identity preservation using identity-based rewards. SHIFT~\citep{Ye2026SHIFTMA} introduces pixel-motion rewards for instantaneous and long-term motion consistency within a hybrid supervised and advantage-weighted fine-tuning framework. Our work adopts DiffusionNFT as the optimization framework and designs a dynamic subject consistency reward that combines local and global signals, substantially improving subject consistency in generated videos, with gains further validated on long videos.

\section{DynSC-Eval}
\label{sec:dynsc-eval}
In this section, we present our evaluation framework. Section~\ref{sec:measurement} describes the evaluation pipeline and defines six metrics covering local and global consistency. Section~\ref{sec:synthetic exp} presents controlled synthetic experiments with deliberately injected inconsistency events, validating the effectiveness of our metrics. Section~\ref{sec:benchmark_analysis} evaluates the reliability and discriminative power of the metrics across diverse video models, demonstrating advantages over conventional metrics such as VBench~\citep{Huang2023VBenchCB}.

\subsection{Measurement}
\label{sec:measurement}

\begin{table}[t]
\centering
\small
\caption{Summary of DynSC-Eval metrics.}
\label{tab:consbench_metrics}
\scalebox{0.8}{
\begin{tabular}{lll}
\toprule
\textbf{Dimension} & \textbf{Metric} & \textbf{Measured Property} \\
\midrule
Local
& Spike Rate (SR)
& Fraction of subjects with inconsistency events \\
Local
& Events/100 Track-s (ER)
& Frequency of inconsistency events \\
Local
& Mean Adjacent Drift (MAD)
& Track-averaged mean adjacent appearance distance \\
\midrule
Global
& Endpoint Drift (ED)
& Start-to-end identity shift \\
Global
& Mean Reference Drift (MRD)
& Average long-term identity drift \\
Global
& Worst-frame Drift (WFD)
& Maximum identity deviation \\
\bottomrule
\end{tabular}
}
\end{table}

\subsubsection{Trajectory Extraction}
We evaluate temporal consistency over \textit{dynamic subject sets}, accounting for all valid subjects throughout the video and construct an explicit trajectory for each target subject.
Formally, given a video $\sV=\{I_t\}_{t=1}^{T}$, we detect target subjects and use SAM2~\citep{Ravi2024SAM2S} to segment and track their visible regions.
Newly observed objects are assigned new trajectories.
After validity filtering, we obtain a trajectory set
$\sT=\{\mathcal{T}_i\}_{i=1}^{N}$, where $N$ is the total number of valid object trajectories across the video.
Each trajectory is represented as
$\mathcal{T}_i = \{(t,x_{i,t}) \mid t \in \mathcal{I}_i\},$
where $\mathcal{I}_i \subseteq \{1,\ldots,T\}$ contains the frames with valid visible observations of object $i$, and $x_{i,t}$ denotes its segmented object crop.
The valid subject set at frame $t$ is therefore defined as:
\begin{equation}
    \mathcal{S}_t = \{i \mid t \in \mathcal{I}_i\},
\end{equation}
which evolves as subjects enter, leave, or become occluded.
Evaluation covers trajectories across $\bigcup_{t=1}^{T}\mathcal{S}_t$.
Each trajectory serves as the basic unit for the subsequent consistency metrics.

For each observation $x_{i,t}$, we extract its visual representation using a pre-trained DINOv3~\citep{Simoni2025DINOv3} encoder $\phi(\cdot)$ and the representation $\boldsymbol{f}_{i,t} = \phi(x_{i,t}).$
All embeddings are $\ell_2$ normalized.
For two observations of the same subject at frames $t,s \in \mathcal{I}_i$, we define their appearance distance as
\begin{equation}
    d(\boldsymbol{f}_{i,t},\boldsymbol{f}_{i,s})
    =
    1-\boldsymbol{f}_{i,t}^{\top}\boldsymbol{f}_{i,s}.
    \label{eq:dino_distance}
\end{equation}
To improve evaluation robustness, we introduce two stabilization strategies: \textit{temporal smoothing} and \textit{trajectory boundary truncation}. Details are provided in Appendix~\ref{a_sec:robust}.

\subsubsection{Local Consistency}
\label{sec:local consistency}
Local consistency measures both \textbf{discrete inconsistency events} and \textbf{continuous short-term appearance variation}. Inconsistency events can be grouped into three categories: (i) \emph{Sudden appearance}: a trajectory begins away from the temporal and spatial boundaries of the video, while the corresponding region in the preceding frame contains no compatible object; (ii) \emph{Sudden disappearance}: a trajectory terminates within the valid region, and no compatible object remains in the corresponding region of the subsequent frame; (iii) \emph{Severe appearance 
mutation}: an existing object undergoes an abrupt deformation or identity change. 
The first two event types are identified from detected trajectory onsets and terminations, subject to the above spatial and temporal boundary and compatibility constraints. For the third type, a transition is classified as a severe appearance mutation if the appearance distance is larger than $\tau$,
where $\tau\in(0,1)$ is a conservative threshold for capturing large and visually salient changes. In the following, we describe the three local consistency metrics and formal definitions are provided in the Appendix~\ref{a_sec:local consistency}

\textbf{Spike Rate (SR):}
The fraction of valid trajectories containing at least one inconsistency event.

\textbf{Events per 100 Track-seconds (ER):}
The number of inconsistency events per 100 seconds of valid trajectory duration, capturing repeated failures.

\textbf{Mean Adjacent Drift (MAD):}
The appearance distance between consecutive temporally smoothed observations, averaged within each valid trajectory and then across trajectories, capturing short-term variation even below the event threshold.

Together, these metrics characterize local consistency through failure prevalence, event frequency, and short-term appearance drift.

\begin{figure}[t]
    \centering
    \includegraphics[width=\textwidth]{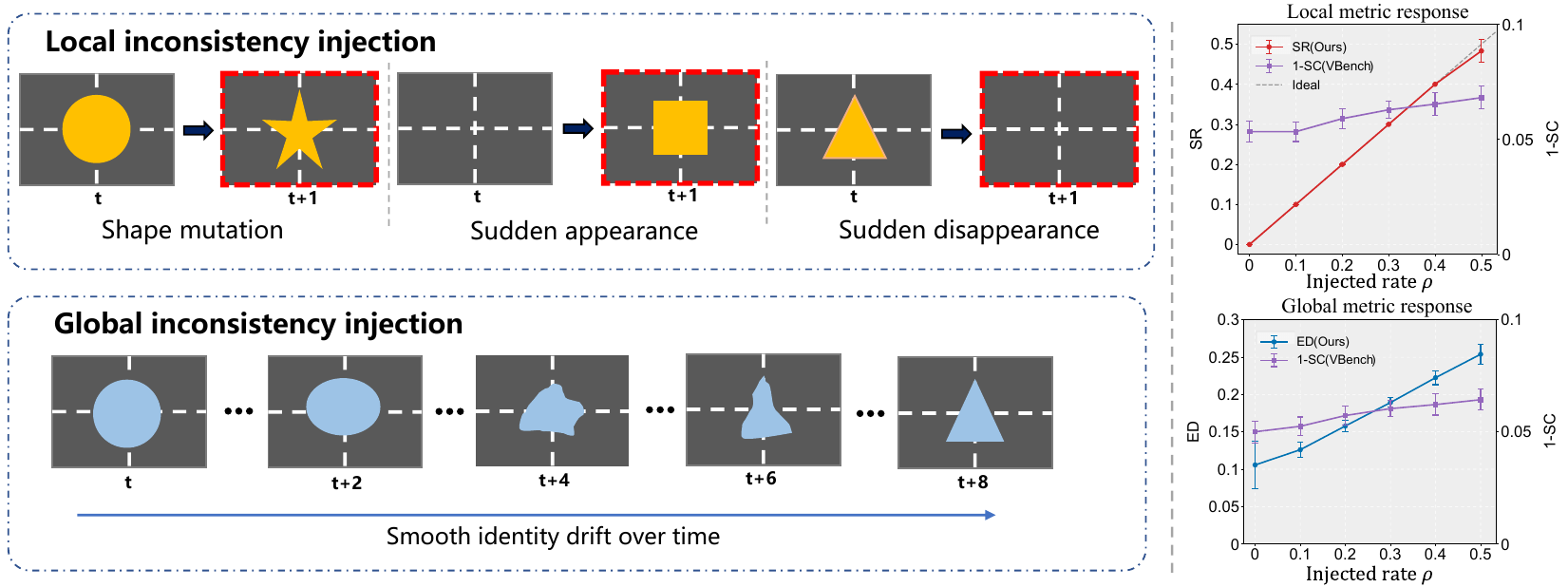}
    \caption{\textbf{Controlled synthetic experiments with actively injected inconsistencies.} We inject local inconsistencies (shape mutation, sudden appearance, sudden disappearance) and global inconsistencies (smooth identity drift) to evaluate metric sensitivity. As the corruption rate increases, our metrics (SR and ED) show a clear response, while VBench (1-SC) largely fails to capture these failures.}
    \label{fig:sythetic}
\end{figure}
\subsubsection{Global Consistency}

Global consistency evaluates whether an object preserves its initially established identity throughout its trajectory. Since global consistency requires a sufficiently long temporal span to establish a reliable reference appearance, we only compute global consistency metrics for trajectories containing at least 16 frames. Trajectories shorter than 16 frames are evaluated only using the local consistency metrics. For each trajectory, we construct a robust identity reference by averaging and normalizing the embeddings of its first $q$ valid observations. An endpoint embedding is similarly obtained.  We next describe the three global consistency metrics, with formal definitions provided in the Appendix~\ref{a_sec:global}

\textbf{Endpoint Drift (ED):}
The appearance difference between the start and end of a trajectory. It computes the cosine distance between the reference and endpoint embeddings, averaged across globally eligible trajectories.

\textbf{Mean Reference Drift (MRD):}
To capture identity variation throughout a trajectory, Mean Reference Drift compares each valid observation against the initial reference. These distances are averaged within each trajectory and then across globally eligible trajectories.

\textbf{Worst-frame Drift (WFD):}
The most severe identity deviation along each trajectory. It takes the maximum distance from the initial reference over valid observations, then averages these maxima across globally eligible trajectories.

The three global metrics respectively characterize final-state identity preservation, average long-term drift, and worst-case deviation. Lower values indicate stronger temporal consistency.

\begin{table*}[t]
\centering
\small
\caption{\textbf{Subject consistency evaluation across diverse models.}
DynSC-Eval metrics are reported to four significant figures (lower is better).
VBench scores are reported on a 0--100 scale (higher is better):
SC denotes Subject Consistency, Dyn denotes Dynamic Degree,
QS denotes the Quality Score aggregated over seven quality dimensions.
Best results within each group are in bold.}
\label{tab:consbench_main}
\resizebox{\linewidth}{!}{%
\begin{tabular}{l*{9}{c}}
\toprule
&
\multicolumn{6}{c}{\textit{DynSC-Eval Metrics}}
&
\multicolumn{3}{c}{\multirow{2}{*}{\textit{VBench Metrics}}}
\\
\cmidrule(lr){2-7}
&
\multicolumn{3}{c}{\textit{Local Consistency}}
&
\multicolumn{3}{c}{\textit{Global Consistency}}
&
\multicolumn{3}{c}{}
\\
\cmidrule(lr){2-4}
\cmidrule(lr){5-7}
\cmidrule(lr){8-10}
\textbf{Model}
& \textbf{SR}$\downarrow$
& \textbf{ER}$\downarrow$
& \textbf{MAD}$\downarrow$
& \textbf{ED}$\downarrow$
& \textbf{MRD}$\downarrow$
& \textbf{WFD}$\downarrow$
& \textbf{SC}$\uparrow$
& \textbf{Dyn}$\uparrow$
& \textbf{QS}$\uparrow$
\\
\midrule

\multicolumn{10}{c}{\textit{Open-source Models (5s)}}\\
\midrule

LTX-Video-2B~\citep{HaCohen2024LTXVideoRV}
& 0.3628 & 28.39 & 0.2629
& 0.3665 & 0.3475 & 0.5474
& 84.58 & 96.00 & 73.82
\\

Wan-2.1-1.3B~\citep{Wang2025WanOA}
& 0.2530 & 10.96 & 0.1757
& 0.3384 & 0.2717 & 0.5041
& 88.09 & 96.00 & 77.09
\\

SANA-2B~\citep{Chen2025SANAVideoEV}
& \textbf{0.1932} & \textbf{7.214} & \textbf{0.1224}
& \textbf{0.3035} & \textbf{0.2081} & \textbf{0.4108}
& \textbf{94.10} & \textbf{99.00} & \textbf{82.75}
\\

Wan-2.1-14B~\citep{Wang2025WanOA}
& 0.1959 & 8.342 & 0.1510
& 0.3065 & 0.2409 & 0.4564
& 92.63 & 94.00 & 79.44
\\

HunyuanVideo-13B~\citep{Kong2024HunyuanVideoAS}
& 0.2583 & 11.95 & 0.1582
& 0.3320 & 0.2550 & 0.4713
& 91.14 & 92.00 & 82.27
\\

\midrule
\multicolumn{10}{c}{\textit{Commercial Models (5s)}}\\
\midrule

MiniMax H3~\citep{minimax2026h3}
& 0.1378 & \textbf{4.427} & \textbf{0.09600}
& 0.2665 & 0.1929 & 0.4042
& \textbf{94.65} & 93.00 & \textbf{84.54}
\\

Seedance 2.0~\citep{Seedance2026Seedance2A}
& \textbf{0.1320} & 4.865 & 0.1012
& 0.2517 & \textbf{0.1841} & \textbf{0.4009}
& 93.92 & 92.00 & 83.56
\\

Kling 3.0~\citep{kling2026v3}
& 0.1414 & 4.954 & 0.1082
& \textbf{0.2492} & 0.1854 & 0.4010
& 94.52 & \textbf{99.00} & 83.13
\\

\midrule
\multicolumn{10}{c}{\textit{Commercial Models (15s)}}\\
\midrule

MiniMax H3~\citep{minimax2026h3}
& 0.1906 & 4.707 & 0.1069
& 0.3343 & 0.2328 & 0.4875
& 92.73 & 92.00 & \textbf{83.74}
\\

Seedance 2.0~\citep{Seedance2026Seedance2A}
& \textbf{0.1634} & 4.257 & 0.1066
& 0.3308 & 0.2296 & 0.4754
& 92.07 & \textbf{97.00} & 83.70
\\

Kling 3.0~\citep{kling2026v3}
& 0.1694 & \textbf{3.677} & \textbf{0.1011}
& \textbf{0.2997} & \textbf{0.2242} & \textbf{0.4662}
& \textbf{93.56} & 93.00 & 83.00
\\

\midrule
\multicolumn{10}{c}{\textit{Streaming Models (60s)}}\\
\midrule
LongLive~\citep{Yang2025LongLiveRI}
& 0.4647 & \textbf{8.356} & \textbf{0.1236}
& 0.4497 & 0.3092 & 0.5988
& \textbf{97.98} & 84.00 & \textbf{82.73}
\\

Helios~\citep{Yuan2026HeliosRR}
& \textbf{0.2870} & 9.066 & 0.1414
& \textbf{0.3704} & \textbf{0.2748} & \textbf{0.5084}
& 94.33 & 96.00 & 77.62
\\

LingBot-World 2.0~\citep{Gao2026InfiniteWW}
& 0.3350 & 10.97 & 0.1424
& 0.4761 & 0.3172 & 0.6403
& 94.16 & \textbf{99.00} & 81.36
\\

\bottomrule
\end{tabular}%
}
\end{table*}

\subsection{Controlled synthetic experiments}
\label{sec:synthetic exp}

To assess the discriminative ability of consistency metrics, we construct a controlled synthetic video experiment with explicitly injected local and global inconsistencies. Each video lasts 30s at a resolution of $832\times480$ and a frame rate of 16 FPS, containing 20 geometric objects that sequentially enter the scene. Their initial shapes are sampled from circles, triangles, squares, and pentagons. Each object moves smoothly along one of two predefined disjoint paths and remains in the scene for 10s.

Fig.~\ref{fig:sythetic}(left) illustrates our inconsistency injection scheme. Local inconsistency simulates sudden appearance or disappearance and a brief appearance mutation:
  the object’s contour becomes a five-pointed star for 8 frames at midlife before returning to its original shape. Global inconsistency simulates persistent shape drift: each object smoothly transitions to a target shape and retains it until leaving the scene.
 Each object undergoes one transition according to the fixed mapping
$\text{circle}\rightarrow\text{triangle}$,
$\text{triangle}\rightarrow\text{square}$,
$\text{square}\rightarrow\text{pentagon}$, and
$\text{pentagon}\rightarrow\text{circle}$.
We interpolate radially between the source and target contours. The  implementation details are provided in Appendix~\ref{app:synthetic-details}.
Under this setting, restricting evaluation to subjects visible in the initial frames can miss many injected inconsistencies. Initial frames may contain no eligible subjects, or the initially visible subjects may remain consistent while inconsistencies occur among later arrivals, demonstrating its limitation. We next assess the sensitivity of DynSC-Eval and compare it with holistic frame-level assessment.

Let $\rho$ denote the proportion of objects with injected inconsistencies. Starting from clean videos ($\rho=0$), we progressively increase $\rho$ and examine the resulting metric responses across three random seeds. For evaluation, we use Grounding DINO~\citep{Liu2023GroundingDM} and SAM2 for automatic object detection and tracking, and DINOv3 features for consistency measurement. 
Fig.~\ref{fig:sythetic}(right) shows the representative results when local and global inconsistencies are jointly injected. SR closely follows the ideal relation $\mathrm{SR}=\rho$, with $R^2=0.995$, while ED increases clearly with $\rho$. In contrast, VBench Subject Consistency (SC) remains nearly unchanged. These results demonstrate the sensitivity of DynSC-Eval to inconsistent subjects. We further isolate the two perturbation types and observe that injecting only local inconsistencies produces little change in ED and MRD, while injecting only global inconsistencies has limited effects on local metrics, demonstrating the complementary roles of our metrics. Results for all six metrics and the isolated ablations are provided in Appendix~\ref{app:synthetic-results}.
\begin{figure}[tb]
    \centering
    \includegraphics[width=0.9\textwidth]{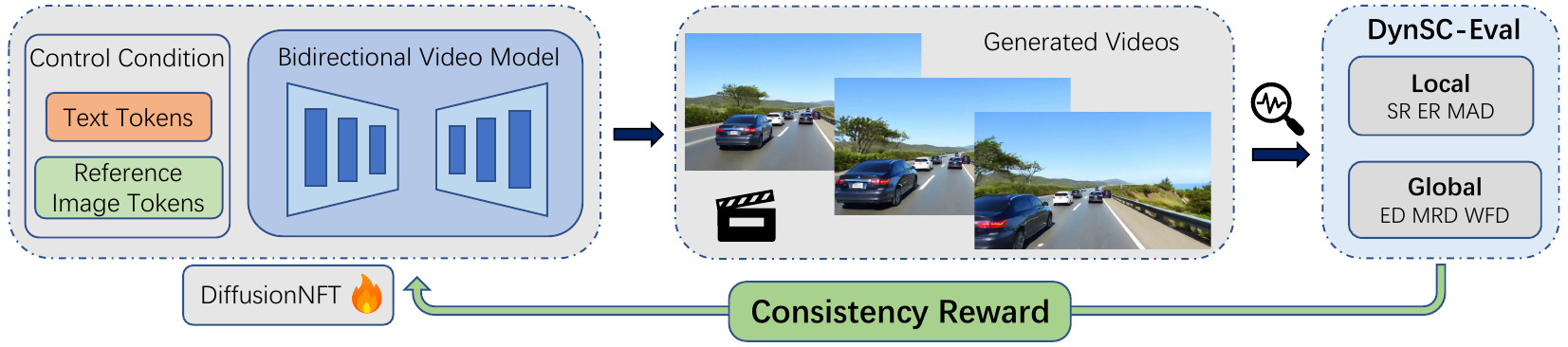}
    \caption{\textbf{Overview of consistency reward optimization framework.} Given text or reference-image conditioning, the video model generates videos that DynSC-Eval assesses through object-level local and global consistency metrics. These metrics are combined into a consistency reward for DiffusionNFT optimization, forming a feedback loop to improve temporal consistency. The framework applies to both T2V and I2V generation across short and long durations.}
    \label{fig:Framework}
\end{figure}
\subsection{Evaluation}
\label{sec:benchmark_analysis}

We construct 100 prompts covering animals, pedestrians, football matches, and autonomous driving, with 25 prompts per scenario. Prompt construction details are provided in Appendix~\ref{app:prompt-construction}. We set the  severe appearance mutation detection threshold to $\tau=0.35$ for general evaluation. We evaluate five open-source models at 5s, three commercial models at both 5s and 15s, and three streaming models at 60s. Table~\ref{tab:consbench_main} shows the detailed evaluation results.

\textbf{Longer videos pose greater consistency challenges.}
For the same commercial models, increasing duration from 5s to 15s raises the mean ED, MRD, and WFD by $25.7\%$, $22.1\%$, and $18.5\%$, respectively, compared with only $3.0\%$ for MAD. SR also increases for all three models, indicating that more trajectories experience severe inconsistencies over longer horizons. The 60s streaming models exhibit substantial inconsistency, with SR ranging from $0.2870$ to $0.4647$, highlighting the remaining challenges in maintaining subject consistency in long videos.

\textbf{DynSC-Eval reveals differences obscured by conventional metrics.}
At 5s, SANA-2B achieves a VBench SC of $94.10$, comparable to commercial models ($93.92$--$94.65$), despite exhibiting noticeably poorer subject consistency under visual inspection. In contrast, DynSC-Eval clearly separates SANA-2B from the commercial models, with SANA-2B performing worse on all six metrics. A similar discrepancy becomes even more evident for long-video generation. The 60s streaming models achieve VBench SC scores comparable to commercial models at 15s, which is clearly inconsistent with visual perception, while DynSC-Eval reveals the differences. These results demonstrate that DynSC-Eval better reflects the inconsistency patterns and provides stronger discrimination across models.

\textbf{DynSC-Eval distinguishes complementary consistency capabilities.}
Seedance 2.0 achieves the lowest SR among commercial models at both durations, whereas Kling 3.0 leads on the remaining five DynSC-Eval metrics at 15s. Among streaming models, LongLive achieves the lowest ER ($8.356$) and MAD ($0.1236$), while Helios performs best on SR ($0.2870$) and all three global metrics. These contrasting profiles demonstrate the value of separately measuring the prevalence of affected subjects, event frequency, short-term variation, and long-term drift.

\section{From Consistency Evaluation to Model Optimization}
\label{sec:optimization}

\subsection{testbed}
\label{sec:testbed}
We next design consistency rewards that directly guide model optimization. We adopt autonomous driving as a representative testbed because driving videos contain multiple persistent vehicles undergoing continuous motion and frequent interactions, which provides abundant object trajectories for exposing both abrupt local failures and accumulated global drift.
We establish two baselines based on Wan-2.1-1.3B~\citep{Wang2025WanOA} and SANA-2B~\citep{Chen2025SANAVideoEV}. We sample real-world videos from OpenDV and divide them into 5s clips, obtaining approximately 100 hours of data. Each clip is automatically captioned using Qwen3-VL-8B~\citep{Bai2025Qwen3VLTR} to construct text--video pairs. Both models are fine-tuned on the same dataset to obtain the baselines. Refer to Appendix~\ref{app:5s-finetune} for details.

\begin{figure}[tb]
    \centering
    \includegraphics[width=\textwidth]{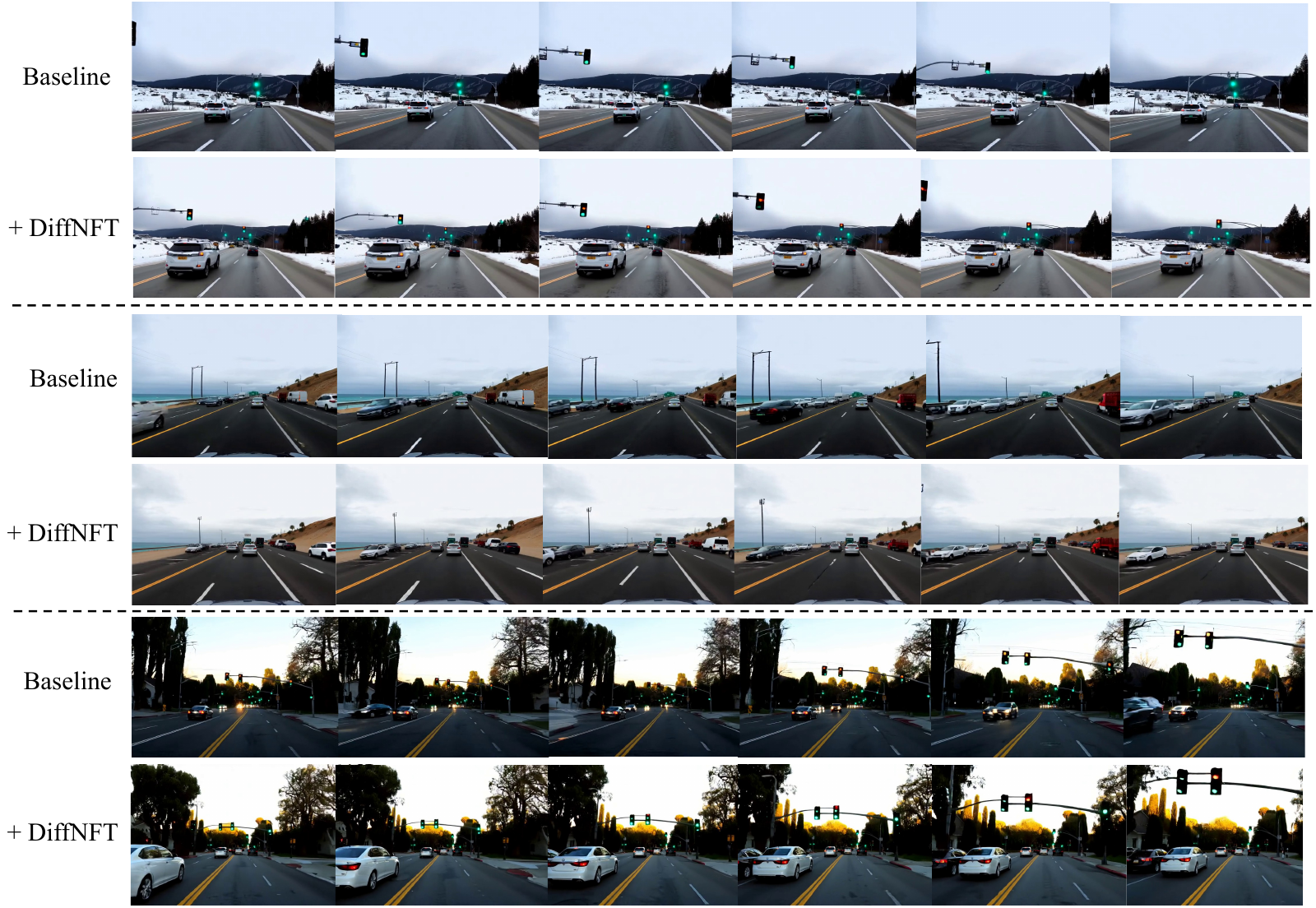}
    \caption{\textbf{Qualitative comparison on Wan-2.1-1.3B.} We compare the baseline with our method after DiffusionNFT. Top: our method eliminates the drift of a distant car in later frames. Middle: the parked car on the left maintains a more stable shape. Bottom: the interpenetration and deformation artifact in the fifth snapshot is removed. Overall, our method produces visually more stable and temporally consistent videos. Additional qualitative results are provided  on the \href{https://dynsc-paper.pages.dev/}{project website}.}
    \label{fig:example}
\end{figure}

\subsection{Method}
\label{sec:consistency_rl}
After establishing the baseline model, we design online rewards from DynSC-Eval metrics and optimize consistency with DiffusionNFT~\citep{Zheng2025DiffusionNFTOD} to further improve subject consistency of these models. Fig.~\ref{fig:Framework} summarizes our consistency optimization framework.

For online optimization, we select four metrics: beta-smoothed SR, MAD, WFD, and MRD, covering both local and global consistency. The consistency reward is defined as
\begin{equation}
R_{\mathrm{cons}} =
-\left(
0.30 C_{\mathrm{spike}}
+
0.30 C_{\mathrm{adj}}
+
0.30 C_{\mathrm{worst}}
+
0.10 C_{\mathrm{ref}}
\right),
\label{eq:cons_reward}
\end{equation}
where each $C$ denotes the normalized inconsistency cost of the corresponding DynSC-Eval metric. A trivial solution would be to generate videos without any target objects. To discourage this behavior, rollouts containing no qualified object trajectories receive a reward of $-10$.

We freeze the base diffusion transformer and optimize LoRA adapters using DiffusionNFT. An EMA-updated old policy generates grouped rollouts, whose rewards are standardized within each prompt group. A frozen reference policy provides regularization to improve consistency while preserving generative capabilities.
Training details are provided in Appendix~\ref{app:post_training}.

\subsection{Experiments}
\label{sec:t2v_results}
We use Qwen3-8B~\citep{Yang2025Qwen3TR} to construct 300 multi-vehicle prompts and independently train the 5s Wan-2.1-1.3B and SANA-2B baselines with DiffusionNFT for 75 steps. Given the stricter requirements for subject consistency in autonomous driving, we use a lower threshold of $\tau=0.16$ to increase sensitivity to severe appearance mutation detection. As shown in Table~\ref{tab:rl_5s}, both models improve across all six DynSC-Eval metrics. For Wan-2.1-1.3B, post-training reduces SR, ER and MAD by $\textbf{13.3\%, 22.2\%, 16.6\%}$. Global consistency also improves, with ED, MRD, and WFD decreasing by $\textbf{10.2\%, 11.0\%}$ and $\textbf{9.6\%}$. SANA-2B exhibits consistent improvements: its three local metrics decrease by $\textbf{5.11\%, 5.99\%, 6.65\%}$ while its three global metrics decrease by $\textbf{6.08\%, 5.96\%}$ and $\textbf{4.19\%}$. Meanwhile, VBench metrics indicate our method improves temporal consistency while preserving other capabilities. 
Qualitative results in Fig.~\ref{fig:example} with additional examples on the \href{https://dynsc-paper.pages.dev/}{project website}, corroborate the quantitative findings, demonstrating the effectiveness of our method.

\begin{table*}[t]
\centering
\small
\caption{\textbf{Results of T2V and
I2V models.}
+DiffNFT denotes the results after 75 optimization steps. DynSC-Eval metrics are reported to four significant figures (lower is better).
VBench scores use a 0–100 scale (higher is better). Better DynSC-Eval results are highlighted in bold.}
\label{tab:rl_5s}
\resizebox{\linewidth}{!}{%
\begin{tabular}{ll*{10}{c}}
\toprule
& &
\multicolumn{6}{c}{\textit{DynSC-Eval Metrics}}
&
\multicolumn{4}{c}{\multirow{2}{*}{\textit{VBench Metrics}}}
\\
\cmidrule(lr){3-8}
& &
\multicolumn{3}{c}{\textit{Local Consistency}}
&
\multicolumn{3}{c}{\textit{Global Consistency}}
&
\multicolumn{4}{c}{}
\\
\cmidrule(lr){3-5}
\cmidrule(lr){6-8}
\cmidrule(lr){9-12}
\textbf{Model}
& \textbf{Setting}
& \textbf{SR}$\downarrow$
& \textbf{ER}$\downarrow$
& \textbf{MAD}$\downarrow$
& \textbf{ED}$\downarrow$
& \textbf{MRD}$\downarrow$
& \textbf{WFD}$\downarrow$
& \textbf{Mot}$\uparrow$
& \textbf{Dyn}$\uparrow$
& \textbf{Aes}$\uparrow$
& \textbf{Img}$\uparrow$
\\
\midrule

\multicolumn{12}{c}{\textit{Text-to-Video (T2V)}}\\
\midrule
\rowcolor{gray!10}
\cellcolor{white}
\multirow{2}{*}{Wan-2.1-1.3B~\citep{Wang2025WanOA}}
& Baseline
& 0.2560
& 9.177
& 0.1059
& 0.3837
& 0.2393
& 0.4428
& 98.35
& 92.00
& 50.47
& 71.40
\\
& +DiffNFT
& \textbf{0.2220} 
& \textbf{7.139} 
& \textbf{0.08829} 
& \textbf{0.3446} 
& \textbf{0.2129} 
& \textbf{0.4002} 
& 98.49
& 92.00
& 50.81
& 71.82
\\
\addlinespace
\hline
\addlinespace

\rowcolor{gray!10}
\cellcolor{white}
\multirow{2}{*}{SANA-2B~\citep{Chen2025SANAVideoEV}}
& Baseline
& 0.2835
& 10.50
& 0.1313
& 0.3979
& 0.2733
& 0.4734
& 98.84
& 91.00
& 53.83
& 71.92
\\
& +DiffNFT
& \textbf{0.2691}
& \textbf{9.868}
& \textbf{0.1225}
& \textbf{0.3737} 
& \textbf{0.2570} 
& \textbf{0.4536} 
& 98.94
& 89.00
& 53.97
& 72.19
\\

\midrule
\multicolumn{12}{c}{\textit{Image-to-Video (I2V)}}\\
\midrule
\rowcolor{gray!10}
\cellcolor{white}
\multirow{2}{*}{ReSim~\citep{Yang2025ReSimRW}}
& Baseline
& 0.2902
& 12.87
& 0.1170
& 0.3170
& 0.2075
& 0.3981
& 98.33
& 91.00
& 50.45
& 71.70
\\
& +DiffNFT
& \textbf{0.2758} 
& \textbf{11.55} 
& \textbf{0.1116} 
& \textbf{0.3068}
& \textbf{0.2013} 
& \textbf{0.3853} 
& 98.44
& 91.00
& 50.44
& 71.26
\\

\bottomrule
\end{tabular}%
}
\end{table*}

\subsection{Generalization to Image-to-Video Generation}
\label{sec:i2v_results}

We next investigate whether our optimization framework can extend to I2V generation. We adopt ReSim~\citep{Yang2025ReSimRW}, an open-source autonomous-driving model at 4.9s, and apply DiffusionNFT to train it with the same 300 prompts. As shown in Table~\ref{tab:rl_5s}, our method improves all six DynSC-Eval metrics, with average reductions of $\textbf{6.61\%}$ and $\textbf{3.14\%}$ for local and global inconsistency metrics, respectively. These results extend our findings beyond T2V generation and support the applicability of the proposed feedback rewards across different generation paradigms.

\subsection{Scaling Consistency Optimization to Long-video Generation}
\label{sec:long_video}

Long-video generation presents a greater challenge to temporal consistency, as object states must remain coherent over longer durations and errors can accumulate over time. Building on the consistency improvements achieved for 5s videos, we next investigate how to improve temporal consistency as generation scales to longer horizons. We adopt a curriculum learning strategy to progressively extend the driving model from 5s to 10s and then 30s. Specifically, we fine-tune the 5s Wan-2.1-1.3B on an additional 100 hours of 10s OpenDV clips, followed by 200 hours of 30s clips, with no overlap among the training subsets. Details are provided in Appendix~\ref{app:curriculum-long}.

We then apply the optimization framework introduced in Sec.~\ref{sec:consistency_rl} to the full-attention long-video models. Tab.~\ref{tab:rl_long} reports the results for 10s and 30s models after 100 training steps, evaluated on fixed sets of 100 and 50 videos, respectively. Our optimization framework consistently improves both local and global consistency. Specifically, for the 10s model, ER, MAD, and SR decrease by \textbf{22.09\%}, \textbf{15.72\%}, and \textbf{12.81\%}, while global MRD and WFD decrease by \textbf{6.20\%} and \textbf{3.69\%}. For the 30s model, the same procedure improves all six metrics: SR, ER, and MAD decrease by \textbf{4.50\%}, \textbf{9.54\%}, and \textbf{2.47\%} while global ED, MRD and WFD decrease by \textbf{8.94\%}, \textbf{9.12\%} and \textbf{7.19\%}. VBench scores remain broadly comparable to the baseline, indicating that visual quality and motion dynamics are largely preserved. Additional qualitative comparisons are provided  on the \href{https://dynsc-paper.pages.dev/}{project website}, further demonstrating the effectiveness  of our optimization framework at longer generation horizons.

\begin{table*}[t]
\centering
\small
\caption{\textbf{Results of the 10s and 30s  models.}
+DiffNFT denotes the results after 100 optimization steps. DynSC-Eval metrics are reported to four significant figures (lower is better).
VBench scores use a 0--100 scale (higher is better). Better DynSC-Eval results are highlighted in bold.}
\label{tab:rl_long}
\resizebox{0.9\linewidth}{!}{%
\begin{tabular}{l*{10}{c}}
\toprule
&
\multicolumn{6}{c}{\textit{DynSC-Eval Metrics}}
&
\multicolumn{4}{c}{\multirow{2}{*}{\textit{VBench Metrics}}}
\\
\cmidrule(lr){2-7}
&
\multicolumn{3}{c}{\textit{Local Consistency}}
&
\multicolumn{3}{c}{\textit{Global Consistency}}
&
\multicolumn{4}{c}{}
\\
\cmidrule(lr){2-4}
\cmidrule(lr){5-7}
\cmidrule(lr){8-11}
\textbf{Setting}
& \textbf{SR}$\downarrow$
& \textbf{ER}$\downarrow$
& \textbf{MAD}$\downarrow$
& \textbf{ED}$\downarrow$
& \textbf{MRD}$\downarrow$
& \textbf{WFD}$\downarrow$
& \textbf{Mot}$\uparrow$
& \textbf{Dyn}$\uparrow$
& \textbf{Aes}$\uparrow$
& \textbf{Img}$\uparrow$
\\
\midrule
\rowcolor{gray!10}
10s baseline
& 0.3843
& 11.77
& 0.1328
& 0.4927
& 0.3291
& 0.5759
& 98.03
& 98.00
& 50.49
& 72.12
\\

+DiffNFT
& \textbf{0.3350}
& \textbf{9.172}
& \textbf{0.1119}
& \textbf{0.4754}
& \textbf{0.3087}
& \textbf{0.5547}
& 97.96
& 98.00
& 50.04
& 72.86
\\

\midrule
\rowcolor{gray!10}
30s baseline
& 0.4684
& 13.27
& 0.1675
& 0.5588
& 0.3712
& 0.6435
& 96.48
& 98.00
& 48.45
& 67.92
\\

+DiffNFT
& \textbf{0.4473}
& \textbf{12.00}
& \textbf{0.1633}
& \textbf{0.5088}
& \textbf{0.3373}
& \textbf{0.5973}
& 96.89
& 98.00
& 48.85
& 67.45
\\

\bottomrule
\end{tabular}%
}
\end{table*}

\section{Ablation Study}
\label{sec:ablation}
We examine the contributions of local and global consistency rewards on Wan-2.1-1.3B for 5s generation. The Local-only reward uses the SR and MAD components, while the Global-only reward uses MRD and WFD.   
As shown in Tab.~\ref{tab:ablation_reward}, Local-only optimization achieves the lowest MAD and all three global drift metrics, but reduces Dyn from $92.00$ to $85.00$, alongside decreases in Aes and Img. This indicates that Local-only optimization tends to converge to a trivial solution of slowing down motion. In contrast, Global-only optimization preserves model capabilities but its local metric SR even worsens. The combined reward achieves the lowest SR and ER, at $0.2220$ and $7.139$, respectively, while maintaining other capabilities comparable to or above the baseline. These results support the necessity and effectiveness of combining local and global rewards.

\begin{table*}[htbp]
\centering
\small
\caption{\textbf{Ablation of consistency reward components.} All optimized models are evaluated at Step 75 using the same configuration. DynSC-Eval metrics are reported to four significant figures (lower is better). VBench scores use a 0--100 scale (higher is better).}
\label{tab:ablation_reward}
\resizebox{0.9\linewidth}{!}{%
\begin{tabular}{l*{10}{c}}
\toprule
& \multicolumn{6}{c}{\textit{DynSC-Eval Metrics}}
& \multicolumn{4}{c}{\multirow{2}{*}{\textit{VBench Metrics}}}
\\
\cmidrule(lr){2-7}
& \multicolumn{3}{c}{\textit{Local Consistency}}
& \multicolumn{3}{c}{\textit{Global Consistency}}
& \multicolumn{4}{c}{}
\\
\cmidrule(lr){2-4}
\cmidrule(lr){5-7}
\cmidrule(lr){8-11}
Reward
& SR$\downarrow$
& ER$\downarrow$
& MAD$\downarrow$
& ED$\downarrow$
& MRD$\downarrow$
& WFD$\downarrow$
& Mot$\uparrow$
& Dyn$\uparrow$
& Aes$\uparrow$
& Img$\uparrow$
\\
\midrule
\rowcolor{gray!10}
5s baseline
& 0.2560 & 9.177 & 0.1059
& 0.3837 & 0.2393 & 0.4428
& 98.35 & 92.00 & 50.47 & 71.40
\\
\midrule

Local-only
& 0.2443 & 7.549 & 0.08653
& 0.3434 & 0.2110 & 0.3929
& 98.61 & 85.00 & 49.81 & 70.84
\\

Global-only
& 0.2691 & 9.155 & 0.09702
& 0.3583 & 0.2262 & 0.4205
& 98.23 & 91.00 & 50.59 & 73.12
\\
\midrule
\rowcolor{blue!10}
Ours (local+global)
& 0.2220 & 7.139 & 0.08829
& 0.3446 & 0.2129 & 0.4002
& 98.49 & 92.00 & 50.81 & 71.82
\\

\bottomrule
\end{tabular}%
}
\end{table*}

\section{Conclusion and Future work}
\label{sec:conclusion}

We highlight the importance of evaluating subject consistency over dynamic subject sets in long video generation. Accordingly, we introduce \textbf{DynSC-Eval}, which tracks eligible subjects and measures local and global consistency, including inconsistency events detection. 
Synthetic experiments and evaluations across diverse models and durations validate its sensitivity and discriminative power, exposing limitations of existing methods. Using autonomous driving as a testbed, we show that rewards derived from DynSC-Eval  improve subject consistency across T2V and I2V models, with gains extending to 10s and 30s generation while preserving other capabilities.

Future work will extend this evaluation to multi-view generation, jointly assessing temporal continuity and consistency across viewpoints. We also aim to incorporate rewards for broader physical constraints into our optimization framework to improve the plausibility of generated dynamics and interactions.

\bibliography{arxiv}
\bibliographystyle{iclr2027_conference}

\newpage
\appendix
\section{Implementation Details of DynSC}
\label{a_sec:DynSC}

\subsection{Local and Global Temporal Consistency}
\label{a_sec:local_global}

For a continuous single-shot video, temporal consistency requires more than visually smooth frame transitions. A generated subject should remain locally coherent from one moment to the next, while also preserving its identity and appearance over a much longer temporal horizon. Accordingly, we decompose temporal consistency into two dimensions: \emph{local consistency} and \emph{global consistency}.

\paragraph{Local temporal consistency.}
Local consistency measures whether a subject's state evolves continuously over short temporal intervals. For nearby observations along the same trajectory, changes caused by motion should occur smoothly rather than abruptly. Local inconsistency manifests as transient events, such as sudden appearance or disappearance and severe deformation, which should be explicitly detected and penalized. This dimension directly reflects the model's ability to maintain short-term temporal continuity.

\paragraph{Global temporal consistency.}
Global consistency measures whether the generative model preserves a subject's established state and identity throughout its complete trajectory. Despite substantial motion and viewpoint changes, later observations should remain compatible with the same underlying subject. This dimension reflects the model's capacity for long-term memory.

Conventional approaches often collapse these behaviors into a single aggregate score, obscuring their distinct failure mechanisms and providing ambiguous supervision for model optimization. In contrast, a more effective approach is to evaluate local discontinuities and global trajectory drift separately and use the resulting subject-trajectory signals as complementary feedback for optimization.

\subsection{Robustness Techniques}
\label{a_sec:robust}
\paragraph{Temporal smoothing.} Recall that $\boldsymbol{f}_{i,k}$ denotes the $\ell_2$-normalized DINOv3 representation of subject $i$ at frame $k$.
Frame-level visual representations may contain small variations.
To obtain a more stable estimate of short-term appearance, we temporally average every two consecutive embeddings with a stride of two as
$\bar{\boldsymbol{f}}_{i,k}
    =
    \operatorname{Norm}(
    \frac{
    \boldsymbol{f}_{i,2k-1}+\boldsymbol{f}_{i,2k}
    }{2}),
    \label{eq:temporal_pooling}$
where $\operatorname{Norm}(\cdot)$ denotes $\ell_2$ normalization.
Adjacent appearance variation is then measured between consecutive pooled representations:
\begin{equation}
    d^{\mathrm{adj}}_{i,k}
    =
    1-
    \bar{\boldsymbol{f}}_{i,k}^{\top}
    \bar{\boldsymbol{f}}_{i,k+1}.
    \label{eq:adjacent_distance}
\end{equation}

\paragraph{Trajectory Boundary Truncation.}
Track initialization and termination are intrinsically ambiguous because a subject may naturally enter or leave the camera view.
Directly including these observations in feature-based computation may introduce noisy distances or false positive signals. 
Let $s_i$ and $e_i$ denote the first and last frame indices of trajectory $\mathcal{T}_i$. We use a temporal margin of $m=4$ frames. If $s_i>m$, we exclude the first $m$ observations of $\mathcal{T}_i$ from feature-based consistency computation. Similarly, if $e_i\leq T-m$, we exclude its last $m$ observations. Although these boundary observations are omitted when computing appearance distances, their neighboring frames are retained for determining whether the trajectory constitutes an abnormal appearance or disappearance event.

\subsection{Metric Computation}
\label{a_sec:computation}
\subsubsection{Local Consistency}
\label{a_sec:local consistency}
Local consistency measures both \textbf{discrete inconsistency events} and \textbf{continuous short-term appearance variation}. We consider three types of inconsistency events: (i) \emph{Sudden appearance}: a trajectory begins away from the temporal and spatial boundaries of the video, while the corresponding region in the preceding frame contains no compatible subject; (ii) \emph{Sudden disappearance}: a trajectory terminates within the valid region, and no compatible subject remains in the corresponding region of the subsequent frame; (iii) \emph{Severe appearance 
mutation}: an existing subject undergoes an abrupt deformation or identity change. 

The first two event types are identified from detected trajectory onsets and terminations, subject to the above spatial and temporal boundary and compatibility constraints. For the third type, a transition is classified as a severe appearance mutation if
$$
d^{\mathrm{adj}}_{i,k}>\tau,
\label{eq:spike_threshold}
$$
where $\tau\in(0,1)$ is a conservative threshold for capturing large and visually salient changes. 

The complete event set for trajectory $i$ is
$$
\mathcal{E}_i
=
\mathcal{E}^{\mathrm{appear}}_i
\cup
\mathcal{E}^{\mathrm{disappear}}_i
\cup
\mathcal{E}^{\mathrm{mutation}}_i.
$$
Let $N_{\mathrm{local}}$ denote the number of trajectories eligible for local evaluation.

\paragraph{Spike Rate (SR).}
Spike Rate measures the fraction of trajectories containing at least one inconsistency event:
\begin{equation}
\mathrm{SR}
=
\frac{1}{N_{\mathrm{local}}}
\sum_{i=1}^{N_{\mathrm{local}}}
\mathbb{I}\left[|\mathcal{E}_i|>0\right].
\label{eq:spike_rate}
\end{equation}

\paragraph{Events per 100 Track-seconds (ER).}
Spike Rate does not distinguish between trajectories containing one failure and those containing repeated failures. We additionally report the number of inconsistency events per 100 seconds of valid subject trajectories computed as $100\cdot{\sum_i|\mathcal{E}_i|}/{\sum_i L_i}$, where $L_i$ denotes the valid duration of trajectory $i$ in seconds.

\paragraph{Mean Adjacent Drift (MAD).}
Mean Adjacent Drift measures continuous short-term appearance variation, including changes below the event threshold:
\begin{equation}
\mathrm{MAD}
=
\frac{1}{N_{\mathrm{local}}}
\sum_{i=1}^{N_{\mathrm{local}}}
\left(
\frac{1}{K_i-1}
\sum_{k=1}^{K_i-1}
d^{\mathrm{adj}}_{i,k}
\right),
\label{eq:mean_adj}
\end{equation}
where $K_i$ is the number of temporally smoothed observations in trajectory $i$.

\subsubsection{Global Consistency}
\label{a_sec:global}
Global consistency evaluates whether a subject preserves its initially established identity throughout its trajectory. Since global consistency requires a sufficiently long temporal span to establish a reliable reference appearance, we only compute global consistency metrics for trajectories containing at least 16 frames. Trajectories shorter than 16 frames are evaluated only using the local consistency metrics.

Let $\mathcal{R}_i$ denote the first $q=4$ valid observations of trajectory $i$, where we use a short multi-frame window rather than a single frame to obtain a robust identity reference.
The reference embedding is
\begin{equation}
    \boldsymbol{f}^{\mathrm{ref}}_i
    =
    \operatorname{Norm}
    (
        \frac{1}{q}
        \sum_{t \in \mathcal{R}_i}
        \boldsymbol{f}_{i,t}).
    \label{eq:reference_embedding}
\end{equation}
Similarly, let $\mathcal{Q}_i$ denote the final $q$ valid observations, from which we construct
\begin{equation}
    \boldsymbol{f}^{\mathrm{end}}_i
    =
    \operatorname{Norm}
    (
        \frac{1}{q}
        \sum_{t \in \mathcal{Q}_i}
        \boldsymbol{f}_{i,t}
    ).
\end{equation}

\paragraph{Endpoint Drift (ED).}
Endpoint Drift measures the appearance difference between the start and end of a trajectory. We macro-average this distance over all globally eligible trajectories and retain:
\begin{equation}
    \mathrm{ED}
    =
    \frac{1}{N_{\mathrm{global}}}
    \sum_i(1-
    (\boldsymbol{f}^{\mathrm{ref}}_i)^{\top}
    \boldsymbol{f}^{\mathrm{end}}_i).
    \label{eq:endpoint}
\end{equation}

\paragraph{Mean Reference Drift (MRD).}
To capture the full evolution of subject identity rather than only its endpoint, each valid observation is compared against the initial reference $$
    d^{\mathrm{ref}}_{i,t}
    =
    1-
    (\boldsymbol{f}^{\mathrm{ref}}_i)^{\top}
    \boldsymbol{f}_{i,t}.
    \label{eq:reference_distance}
$$
The average long-term deviation of trajectory $i$ is $$
    D^{\mathrm{ref}}_{i,\mathrm{mean}}
    =
    \frac{1}{|\mathcal{V}_i|}
    \sum_{t \in \mathcal{V}_i}
    d^{\mathrm{ref}}_{i,t},$$
where $\mathcal{V}_i$ denotes the valid evaluation frames.
We define
\begin{equation}
    \mathrm{MRD}
    =
    \frac{1}{N_{\mathrm{global}}}
    \sum_i
    D^{\mathrm{ref}}_{i,\mathrm{mean}}.
    \label{eq:mean_ref}
\end{equation}

\paragraph{Worst-frame Drift (WFD).}
Worst-frame Drift captures the most severe identity deviation along each trajectory.
We average the worst-frame deviations across trajectories and define:
\begin{equation}
    \mathrm{WFD}
    =
    \frac{1}{N_{\mathrm{global}}}
    \sum_i
    \max_{t \in \mathcal{V}_i}
    d^{\mathrm{ref}}_{i,t}.
    \label{eq:worst_frame}
\end{equation}

\subsection{Details of the Synthetic Experiment}
\label{app:synthetic-details}

\paragraph{Scene construction.}
We construct paired synthetic videos to examine how consistency
metrics respond to controlled shape perturbations at different
temporal scales.
Each video contains 480 frames at 16 FPS, with a resolution of
$832\times480$.
A total of 20 geometric objects are introduced sequentially along
two predefined motion paths.
Their initial shapes are sampled from circles, triangles, squares,
and pentagons, with radius parameters ranging from 72 to 84 pixels.
Each object has a scheduled lifetime of 10 seconds.
This lifetime includes entering and leaving the image boundary;
the interval during which its contour is visible can therefore
be slightly shorter.
In the following, object age $t$ denotes elapsed time since its
scheduled birth, rather than since its first visible pixel.
Object colors, motion trajectories, birth times, and radius
parameters are shared across paired conditions.

\paragraph{Smooth global perturbations.}
A selected object undergoes one source-to-target shape transition
according to the mapping
$\mathrm{circle}\rightarrow\mathrm{triangle}$,
$\mathrm{triangle}\rightarrow\mathrm{square}$,
$\mathrm{square}\rightarrow\mathrm{pentagon}$, or
$\mathrm{pentagon}\rightarrow\mathrm{circle}$.
It retains its source shape for $t\in[0,2]$, transitions during
$t\in(2,8)$, and retains its target shape for the remainder of
its lifetime.
Let $r_{\mathrm{src}}(\theta)$ and $r_{\mathrm{tgt}}(\theta)$
denote normalized radial profiles of the source and target shapes.
We generate the intermediate contour using
\begin{equation}
\begin{aligned}
r(\theta,t)
&= (1-\alpha(t))r_{\mathrm{src}}(\theta)
 + \alpha(t)r_{\mathrm{tgt}}(\theta),\\
\alpha(t)
&= 3u^2-2u^3,\\
u
&= \operatorname{clip}\!\left(\frac{t-2}{6},0,1\right).
\end{aligned}
\label{eq:synthetic-interpolation}
\end{equation}
For an object with center $\mathbf{c}(t)$ and radius parameter $R$,
the corresponding contour points are
\begin{equation}
\mathbf{p}(\theta,t)
=
\mathbf{c}(t)
+
R\,r(\theta,t)
\begin{bmatrix}
\cos\theta\\
\sin\theta
\end{bmatrix}.
\end{equation}
We sample 360 uniformly spaced angular directions and rasterize
the resulting contour.
The smoothstep interpolation gives zero temporal derivative at
the beginning and end of the transition.
The center trajectory, radius parameter, and color remain fixed
across conditions, while the contour area may change naturally.
All conditions use the same contour renderer, including Clean.

\paragraph{Local inconsistency injection.}
  We assign selected objects as evenly as possible to three
  types of local inconsistency: severe appearance mutations, sudden
  appearances, and sudden disappearances.
  Each selected object receives one perturbation.
  For appearance mutations, we replace the object's current
  contour with a deeply concave five-pointed star at an age
  of approximately 5 seconds.
  The onset is rounded to the nearest native video frame,
  and the replacement lasts exactly 8 frames (0.5 seconds).
  The star has an inner-to-outer radius ratio of 0.1 and
  preserves the object's center, outer radius parameter,
  and color.
  Afterward, the object resumes its unperturbed shape evolution.
  For abrupt appearances, we hide the object during its natural
  entry and reveal it at an age of 3 seconds.
  For abrupt disappearances, we remove the object at an age
  of 6.5 seconds for the remainder of its scheduled lifetime.
  Both events occur inside the image, with onsets mapped to
  the first native frame at or after the prescribed time. 

\paragraph{Corruption rates and paired controls.}
We use three random seeds, 101, 202, and 303, and corruption rates
$\rho\in\{0,0.1,0.2,0.3,0.4,0.5\}$.
The denominator of $\rho$ is always the full set of 20 objects,
so the five nonzero rates correspond to 2, 4, 6, 8, and 10
perturbed objects.
For an object with scheduled birth time $b$, eligibility for
perturbation requires $b\geq0.5$ and $b+10<28$.
These constraints yield 15 eligible objects per seed and ensure
that the complete global transition and subsequent departure
occur within the protected temporal boundaries.
We fix an ordering of eligible objects before evaluation and
select nested prefixes as $\rho$ increases.
The selected set is identical across Local-only, Global-only,
and Local+Global for each seed and rate.

\begin{figure}[tb]
    \centering
    \includegraphics[width=\textwidth]{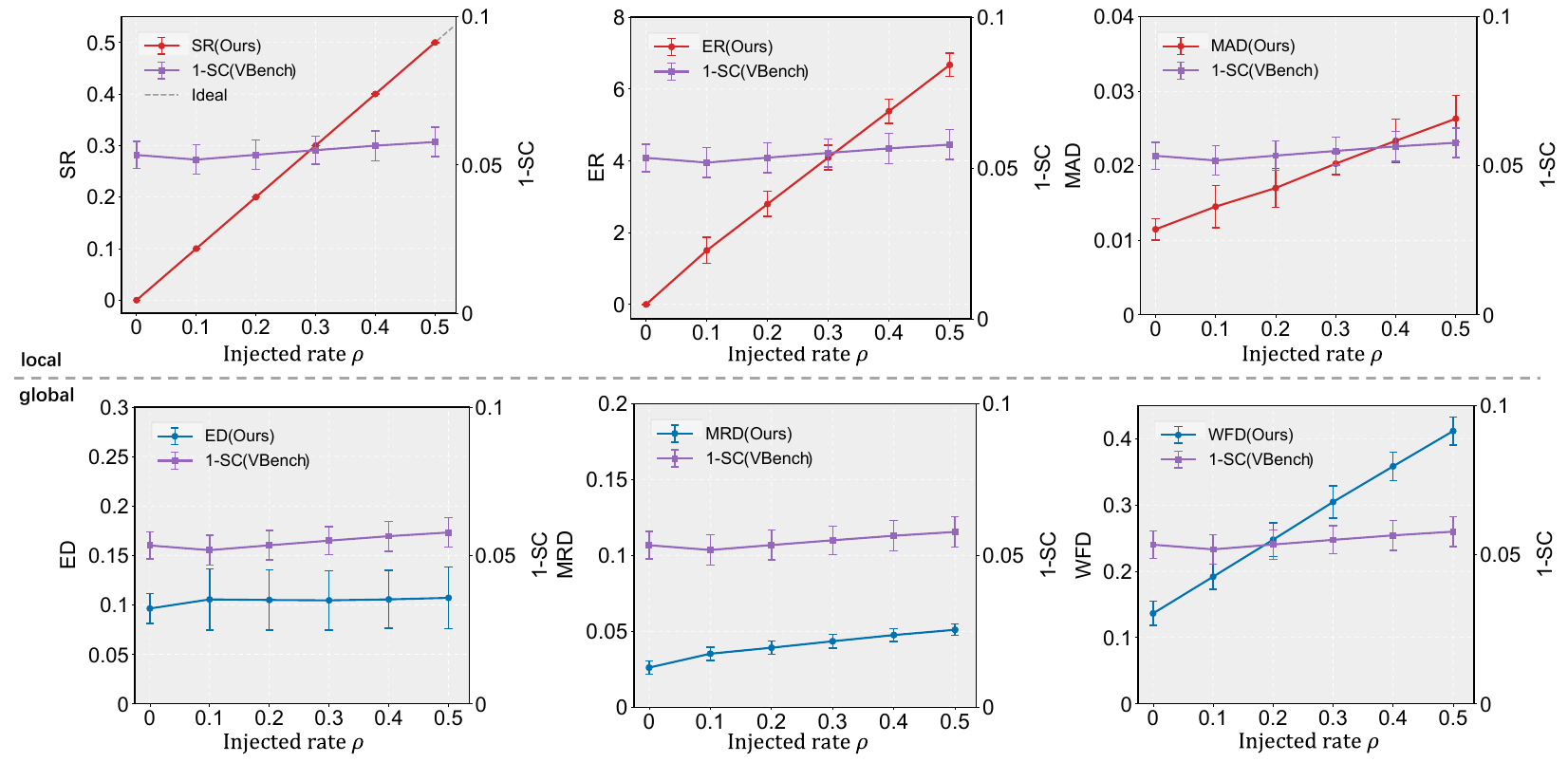}
    \caption{\textbf{  Metric responses to Local-only inconsistency injection. } As the fraction of objects with local inconsistencies                                                    
  increases, SR and ER rise, while ED remains nearly unchanged.     MRD also responds to these temporary deviations from the
  reference appearance.                                   Each point shows the mean over three random seeds, with                                                         error bars indicating one sample standard deviation.
  VBench is shown as $1-\mathrm{SC}$ on the right axis.}
    \label{fig:synthetic_local}
\end{figure}
\begin{figure}[tb]
    \centering
    \includegraphics[width=\textwidth]{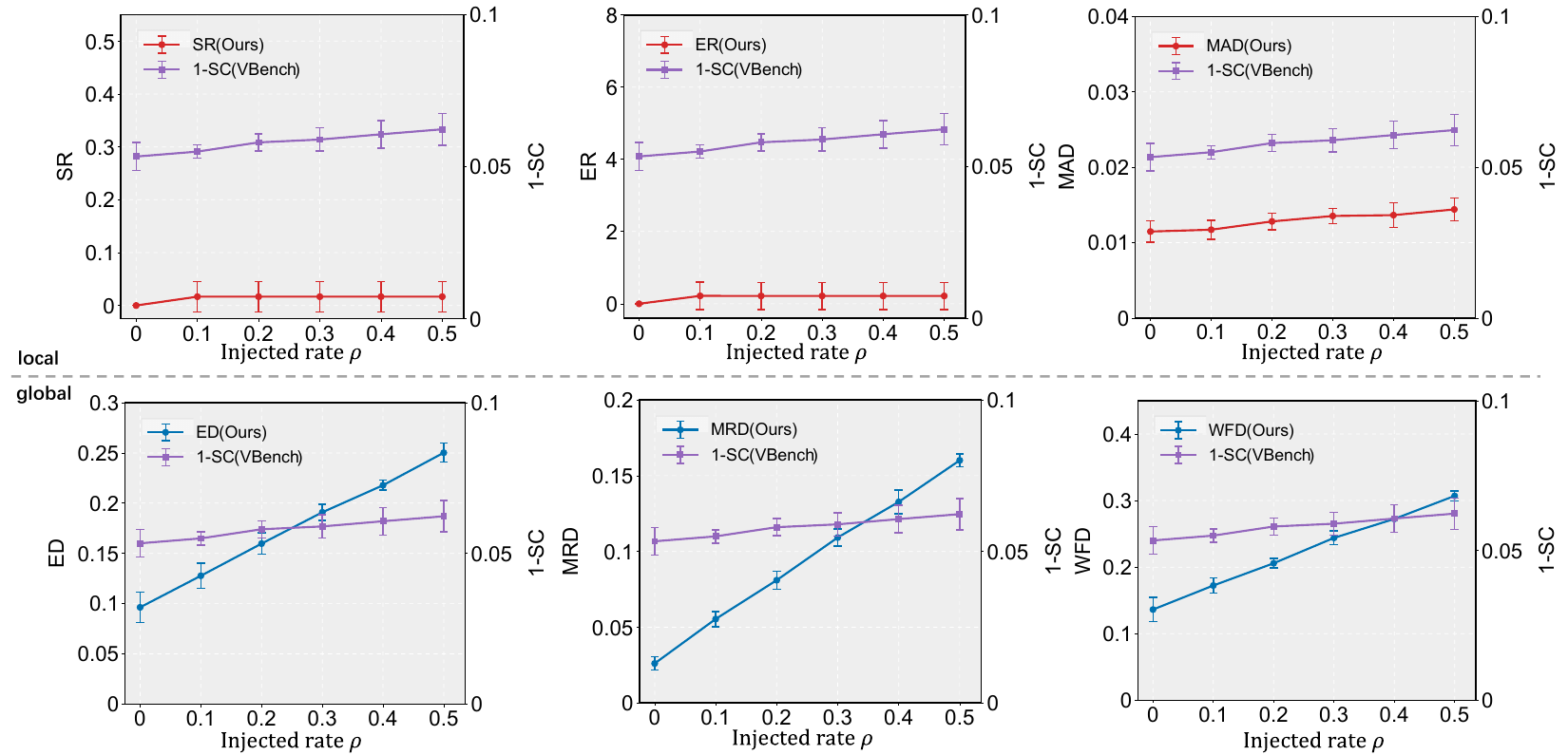}
    \caption{\textbf{Metric responses to Global-only inconsistency injection.} As the fraction of objects undergoing smooth shape transitions
  increases, ED and MRD rise, while SR and ER remain so close to clean
  baseline and  change minimally.
  Each point shows the mean over three random seeds, with
  error bars indicating one sample standard deviation.
  VBench is shown as $1-\mathrm{SC}$ on the right axis.}
    \label{fig:synthetic_global}
\end{figure}
\begin{figure}[tb]
    \centering
    \includegraphics[width=\textwidth]{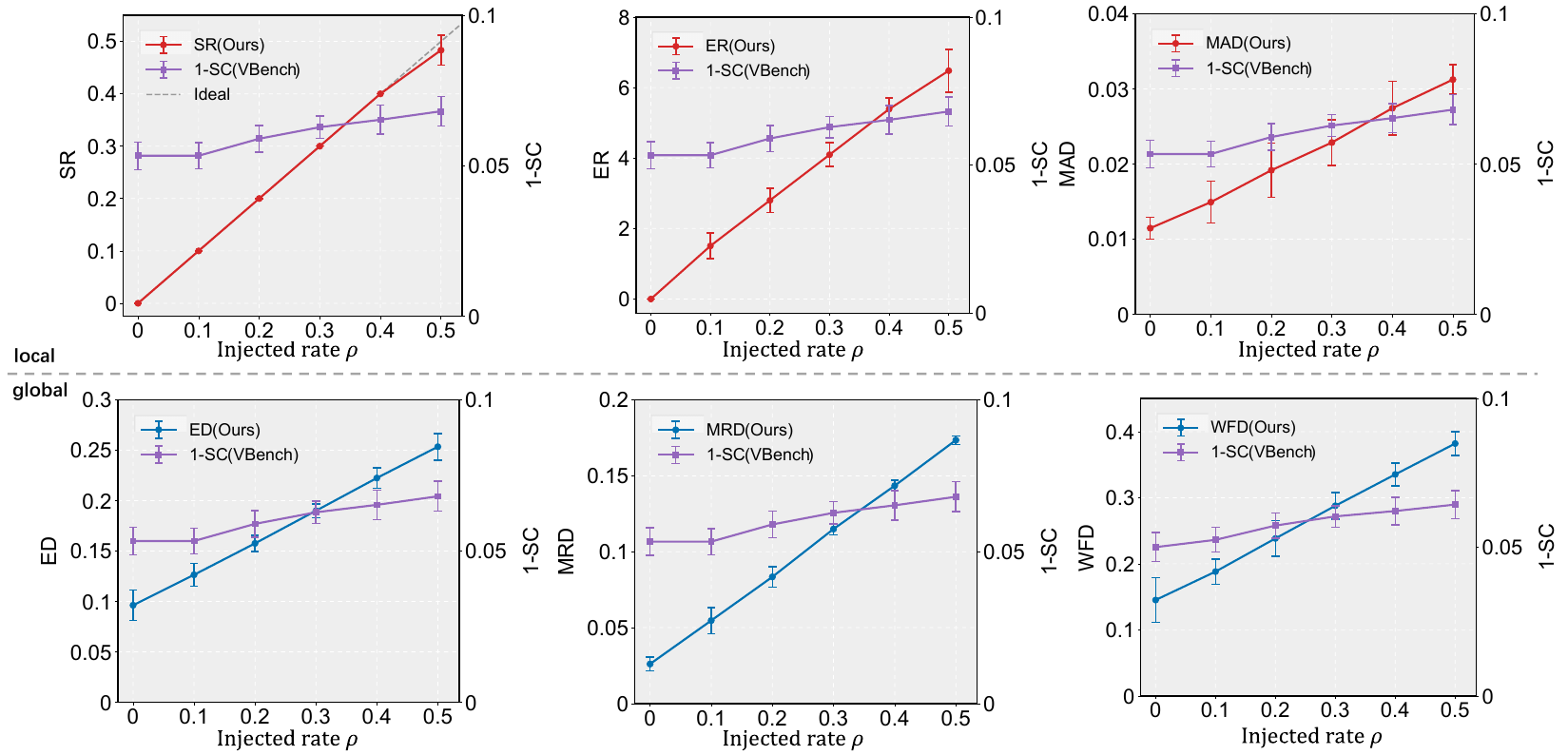}
    \caption{\textbf{Metric responses to joint local and global inconsistency injection.} 
  As the fraction of perturbed objects increases, SR closely follows
  the ideal relation $\mathrm{SR}=\rho$, while ER, ED, and MRD increase.
  Each point shows the mean over three random seeds, with
  error bars indicating one sample standard deviation.
  VBench is shown as $1-\mathrm{SC}$ on the right axis.}
    \label{fig:synthetic_local+global}
\end{figure}
\subsection{Ablation of Local-only and Global-only Injections}
\label{app:synthetic-results}
The isolated conditions clarify the different roles of the
measurements. Fig.~\ref{fig:synthetic_local} shows the Local-only results.
At $\rho=0.5$, averaged over three seeds, Local-only increases
SR from 0 to 0.5000 and ER from 0 to 6.6687 events
per 100 track-seconds.
 ED remains relatively stable, changing from 0.0963 to 0.1071.
Conversely, as shown in Fig~\ref{fig:synthetic_global},  Global-only increases ED from 0.0963 to 0.2502 and MRD from 0.0261 to 0.1601. In contrast,
SR remains very low, with a mean of 0.0167 at every
  nonzero corruption level and
  ER only reaches 0.2173 events per 100 track-seconds at $\rho=0.5$,
  substantially below its response to Local-only injections.
Fig~\ref{fig:synthetic_local+global} demonstrates the results of local and global inconsistency injection. Under joint injection, SR and ER reach 0.4833 and 6.4790,
respectively, while ED and MRD reach 0.2535 and 0.1735.
These observations support the complementary responses of
local metrics and global metrics to transient disruptions.

To quantify agreement between measured SR and the injected
spike rate, we compare the measurements directly with the
identity relation, without fitting a slope or intercept.
Let $y_{s,k}$ be the measured SR for seed $s$ at corruption
rate $\rho_k$.
For Local-only and Local+Global, we compute
\begin{equation}
R^2_{\mathrm{identity}}
=
1-
\frac{\sum_{s,k}(y_{s,k}-\rho_k)^2}
     {\sum_{s,k}(\rho_k-\bar{\rho})^2},
\qquad
\bar{\rho}=0.25.
\label{eq:synthetic-identity-r2}
\end{equation}
The calculation includes all 18 seed--rate measurements per
condition, including the clean baseline, rather than only
the six seed-averaged curve points.
For Local+Global, this gives
  $R^2_{\mathrm{identity}}=0.9952$, with an MAE of 0.0028 and
  an RMSE of 0.0118.
  Local-only yields $R^2_{\mathrm{identity}}=1.0000$,
  with both MAE and RMSE equal to zero. All injected local inconsistency episodes were detected,
  consistent with the intended sensitivity of SR to the injected local inconsistency
    event. The identity relation is not an appropriate target for
  Global-only, whose injected spike rate remains zero.

\subsection{Prompt Construction}
\label{app:prompt-construction}

Our evaluation covers four representative scenarios: animals, pedestrians, football matches, and autonomous driving, with 25 prompts per scenario. For reproducibility, the complete set of evaluation prompts used in all experiments is provided on the \href{https://dynsc-paper.pages.dev/}{project website}.

\subsection{Video Generation Prompts}
\label{app:video-generation-prompts}

The following boxes demonstrate the text prompts associated with the accompanying video examples in Fig.~\ref{fig:failure-mode}. Paragraphs and headings are formatted for readability. 

\begin{consbenchprompt}{Seedance 2.0 \textbar\ 15\,s}
\label{prompt:seedance-2.0}
\textbf{Camera and video style requirements}\par\smallskip
Every generated video must be a realistic single continuous shot. Required: - The entire video must be captured from one continuous camera trajectory. - The camera should maintain temporal continuity throughout the 15-second duration. - The scene should feel like a real-world observation video. Allowed: - smooth camera tracking; - handheld camera movement; - dashcam view; - fixed surveillance camera; - gradual panning; - natural viewpoint changes. Strictly avoid: - cuts; - scene transitions; - camera switching; - replay; - montage; - time jumps; - sudden viewpoint changes.

\medskip
\textbf{Motion requirements}\par\smallskip
The video should contain normal-speed realistic motion. Required: - Subjects should move at natural real-world speed. - Motion should reflect realistic physics and interactions. Strictly avoid: - slow motion; - slow-motion effects; - cinematic slow-motion shots; - time-lapse; - fast-forward effects; - exaggerated motion; - unrealistic acceleration. The generated video should resemble a real continuous recording rather than a cinematic highlight or edited movie sequence.

\medskip
\textbf{Scene specification}\par\smallskip
An elevated stadium camera records a youth football tournament match between a green team and a white team. Twelve young players are visible, including anchor players: a green captain wearing number 10, a green midfielder wearing number 8, a green defender wearing number 2, a white forward wearing number 9, and a white midfielder wearing number 6. Around 3 seconds, players establish their positions and begin moving around the field. Around 6 seconds, the green team performs several short passes while players rotate positions and opponents follow closely. Around 9 seconds, multiple players converge near the ball, briefly overlapping while maintaining individual appearances. Around 12 seconds, players separate into attacking and defensive formations. The final frames preserve jersey numbers, hairstyles, and player identities. Single continuous shot without slow motion.
\end{consbenchprompt}

\begin{consbenchprompt}{MiniMax H3 \textbar\ 15\,s}
\label{prompt:minimax-h3}
\textbf{Camera and video style requirements}\par\smallskip
Every generated video must be a realistic single continuous shot. Required: - The entire video must be captured from one continuous camera trajectory. - The camera should maintain temporal continuity throughout the 15-second duration. - The scene should feel like a real-world observation video. Allowed: - smooth camera tracking; - handheld camera movement; - dashcam view; - fixed surveillance camera; - gradual panning; - natural viewpoint changes. Strictly avoid: - cuts; - scene transitions; - camera switching; - replay; - montage; - time jumps; - sudden viewpoint changes.

\medskip
\textbf{Motion requirements}\par\smallskip
The video should contain normal-speed realistic motion. Required: - Subjects should move at natural real-world speed. - Motion should reflect realistic physics and interactions. Strictly avoid: - slow motion; - slow-motion effects; - cinematic slow-motion shots; - time-lapse; - fast-forward effects; - exaggerated motion; - unrealistic acceleration. The generated video should resemble a real continuous recording rather than a cinematic highlight or edited movie sequence.

\medskip
\textbf{Scene specification}\par\smallskip
An elevated stadium camera records a professional football match between a yellow team and a dark blue team. The scene contains 14 visible players. Five anchor players are a yellow goalkeeper wearing number 1, yellow defender wearing number 4, yellow midfielder wearing number 8, dark blue striker wearing number 10, and dark blue defender wearing number 6. Around 3 seconds, players gather near the penalty area preparing for a corner kick. Around 6 seconds, players begin running and marking opponents as the ball enters play. Around 8 seconds, attackers and defenders move tightly together, causing temporary overlaps and occlusions. Around 11 seconds, players separate and continue tracking the ball. Around 13 seconds, the camera follows the movement while preserving the identity of all anchor players. No replay, no camera cuts.
\end{consbenchprompt}

\begin{consbenchprompt}{Kling 3.0 \textbar\ 15\,s}
\label{prompt:kling-v3}
\textbf{Camera and video style requirements}\par\smallskip
Every generated video must be a realistic single continuous shot. Required: - The entire video must be captured from one continuous camera trajectory. - The camera should maintain temporal continuity throughout the 15-second duration. - The scene should feel like a real-world observation video. Allowed: - smooth camera tracking; - handheld camera movement; - dashcam view; - fixed surveillance camera; - gradual panning; - natural viewpoint changes. Strictly avoid: - cuts; - scene transitions; - camera switching; - replay; - montage; - time jumps; - sudden viewpoint changes.

\medskip
\textbf{Motion requirements}\par\smallskip
The video should contain normal-speed realistic motion. Required: - Subjects should move at natural real-world speed. - Motion should reflect realistic physics and interactions. Strictly avoid: - slow motion; - slow-motion effects; - cinematic slow-motion shots; - time-lapse; - fast-forward effects; - exaggerated motion; - unrealistic acceleration. The generated video should resemble a real continuous recording rather than a cinematic highlight or edited movie sequence.

\medskip
\textbf{Scene specification}\par\smallskip
A realistic dashcam video shows a busy downtown avenue during daytime with a continuous single camera shot from a moving vehicle. The scene contains 12 visible road vehicles, including a red compact sedan, a white SUV, a blue delivery van, a silver city bus, a black pickup truck, a yellow taxi, a gray hatchback, a green motorcycle, a white cargo truck, a dark blue sedan, a silver minivan, and a brown crossover SUV. The red compact sedan, white SUV, blue delivery van, silver city bus, and black pickup truck are anchor vehicles that remain visible throughout the entire 15-second video. Around 3--5 seconds, the yellow taxi accelerates from the right lane and moves ahead of the gray hatchback, while the white SUV gradually slows down due to a traffic light. Around 6--10 seconds, the blue delivery van changes lanes and briefly passes close to the red sedan, causing temporary partial occlusion between the two vehicles while both vehicles remain recognizable. Around 10--13 seconds, the black pickup truck overtakes the silver minivan while the city bus continues straight and the relative positions of the anchor vehicles keep changing naturally. Near the end of the video, all anchor vehicles remain visible with consistent appearance, colors, and identities.
\end{consbenchprompt}

\begin{consbenchprompt}{LongLive \textbar\ 60\,s \textbar\ Example 1}
\label{prompt:longlive-ex1}
Generate one realistic 60-second video in a single continuous, uninterrupted shot at normal speed. Keep 10-15 subjects visible at detector-friendly scale in a wide or medium-wide view. At least five distinctive anchor subjects must remain continuously visible and preserve identity, color, shape, scale, and markings for the full minute. Use one smooth camera trajectory and physically plausible motion. No cuts, transitions, camera switches, replay, montage, time jumps, slow motion, abrupt viewpoint changes, severe occlusion, tiny distant subjects, teleportation, duplication, disappearance, identity swaps, transformations, text overlays, or watermarks.

\medskip
A 60-second handheld walking camera video inside a large outdoor shopping street with stores on both sides. There are 13 visible pedestrians. Five anchor pedestrians include a woman with short blonde hair wearing a blue jacket, a man with a red backpack wearing a black shirt, a woman wearing a pink hat and beige coat, a teenager wearing a green hoodie, and an elderly person wearing a navy sweater with glasses. During 0-10 seconds, pedestrians browse shops and walk along the street. During 10-25 seconds, several pedestrians change directions, one person stops briefly near a store entrance, and some pedestrians pass behind others causing temporary partial occlusion. The camera continues moving forward. Single shot, realistic shopping street activity. During 25-40 seconds, the pedestrians vary walking speed and relative position while the camera changes viewpoint gradually and keeps the five anchors in view. During 40-52 seconds, the five anchor pedestrians take part in a second natural crossing and grouping interaction, briefly overlap, then separate with the same clothing, accessories, faces, and body proportions. During 52-60 seconds, normal walking continues and all five anchor pedestrians remain visible, recognizable, and unchanged through the final frame.
\end{consbenchprompt}

\begin{consbenchprompt}{LongLive \textbar\ 60\,s \textbar\ Example 2}
\label{prompt:longlive-ex2}
Generate one realistic 60-second video in a single continuous, uninterrupted shot at normal speed. Keep 10-15 subjects visible at detector-friendly scale in a wide or medium-wide view. At least five distinctive anchor subjects must remain continuously visible and preserve identity, color, shape, scale, and markings for the full minute. Use one smooth camera trajectory and physically plausible motion. No cuts, transitions, camera switches, replay, montage, time jumps, slow motion, abrupt viewpoint changes, severe occlusion, tiny distant subjects, teleportation, duplication, disappearance, identity swaps, transformations, text overlays, or watermarks.

\medskip
A realistic 60-second single continuous shot. This is association soccer played with a round soccer ball; players wear standard soccer shirts, shorts, socks, and cleats on a regulation soccer pitch. A wide-angle broadcast camera records a professional soccer match in a large stadium with two teams, one wearing red jerseys and the other wearing blue jerseys. At the beginning, 12 players are visible, including five anchor players: a tall red midfielder wearing number 8, a short red winger wearing number 11, a blue defender wearing number 4, a blue midfielder wearing number 6, and a goalkeeper wearing number 1. During 0-10 seconds, both teams begin a midfield transition as players run into attacking and defensive positions. During 10-25 seconds, the red midfielder passes the ball forward while multiple players cross paths near the center circle and briefly overlap. During 25-40 seconds, defenders mark attackers closely while the camera tracks the moving group from the sideline. During 25-40 seconds, players spread out again while maintaining their original jersey colors, numbers, and appearances. No replay, no slow motion, no camera cuts. During 40-52 seconds, the five anchor players join a second passing and defending sequence, cross and briefly overlap with teammates and opponents, then separate with the same jerseys, numbers, positions, and appearances. During 52-60 seconds, play continues at normal speed and all five anchor players remain visible, recognizable, and unchanged through the final frame.
\end{consbenchprompt}

\begin{consbenchprompt}{LingBot-World 2.0 \textbar\ 60\,s \textbar\ Example 1}
\label{prompt:lingbot-world-ex1}
Generate one realistic 60-second video in a single continuous, uninterrupted shot at normal speed. Keep 10-15 subjects visible at detector-friendly scale in a wide or medium-wide view. At least five distinctive anchor subjects must remain continuously visible and preserve identity, color, shape, scale, and markings for the full minute. Use a locked-off static camera with zero translation, zero rotation, and no zoom; only the subjects move, with physically plausible motion. No cuts, transitions, camera switches, replay, montage, time jumps, slow motion, abrupt viewpoint changes, severe occlusion, tiny distant subjects, teleportation, duplication, disappearance, identity swaps, transformations, text overlays, or watermarks.

\medskip
A 60-second fixed wide-angle camera video showing pedestrians through a university campus pathway during afternoon. The scene contains 11 visible pedestrians. Five anchor pedestrians remain throughout: a student wearing a red jacket with a black backpack, a student wearing a white sweater and glasses, a tall person wearing a blue cap and denim jacket, a woman with braided hair carrying a laptop bag, and an older professor wearing a gray coat. During 0-10 seconds, students walk between buildings, some talking in pairs and others walking alone. During 10-25 seconds, two groups pass each other, several pedestrians briefly overlap, and one student walks behind another before becoming visible again. The locked-off camera maintains a stable wide view. Natural campus environment, single continuous shot. During 25-40 seconds, the pedestrians vary walking speed and relative position. During 40-52 seconds, the five anchor pedestrians take part in a second natural crossing and grouping interaction, briefly overlap, then separate with the same clothing, accessories, faces, and body proportions. During 52-60 seconds, normal walking continues and all five anchor pedestrians remain visible, recognizable, and unchanged through the final frame.
\end{consbenchprompt}

\begin{consbenchprompt}{LingBot-World 2.0 \textbar\ 60\,s \textbar\ Example 2}
\label{prompt:lingbot-world-ex2}
Generate one realistic 60-second video in a single continuous, uninterrupted shot at normal speed. Keep 10-15 subjects visible at detector-friendly scale in a wide or medium-wide view. At least five distinctive anchor subjects must remain continuously visible and preserve identity, color, shape, scale, and markings for the full minute. Use a locked-off static camera with zero translation, zero rotation, and no zoom; only the subjects move, with physically plausible motion. No cuts, transitions, camera switches, replay, montage, time jumps, slow motion, abrupt viewpoint changes, severe occlusion, tiny distant subjects, teleportation, duplication, disappearance, identity swaps, transformations, text overlays, or watermarks.

\medskip
A realistic 60-second single continuous shot. A fixed wide-angle view shows a group of twelve dogs running along a wide forest trail in a natural park. The group includes a large white German shepherd with a black saddle marking, a brown Labrador with a red collar, a small cream-colored Corgi with short legs, a gray Australian shepherd with blue eyes, a black-and-white Border Collie, and several other dogs of different breeds. The locked-off camera maintains a stable wide view. During 0-10 seconds, the dogs become clearly separated, revealing their unique fur colors, body sizes, and breed characteristics. The German shepherd remains near the front, while the Corgi moves near the center of the group. During 10-25 seconds, several dogs change positions while running, with some crossing paths and briefly walking beside each other. The Border Collie runs around the Labrador before returning to the main group. The locked-off camera maintains a stable wide view. During 25-40 seconds, the dogs slow down slightly and begin walking together while exploring the forest path. Their fur patterns, body shapes, and individual markings remain unchanged throughout the movement. During 40-52 seconds, the five anchor animals reorganize once more: two cross paths, one is briefly partially overlapped, and all five separate again with the same coats, markings, sizes, and body shapes. During 52-60 seconds, the group continues natural motion and all five anchor animals remain visible, recognizable, and unchanged through the final frame.
\end{consbenchprompt}

\begin{consbenchprompt}{Helios \textbar\ 60\,s \textbar\ Example 1}
\label{prompt:helios-ex1}
Generate one realistic 60-second video in a single continuous, uninterrupted shot at normal speed. Keep 10-15 subjects visible at detector-friendly scale in a wide or medium-wide view. At least five distinctive anchor subjects must remain continuously visible and preserve identity, color, shape, scale, and markings for the full minute. Use one smooth camera trajectory and physically plausible motion. No cuts, transitions, camera switches, replay, montage, time jumps, slow motion, abrupt viewpoint changes, severe occlusion, tiny distant subjects, teleportation, duplication, disappearance, identity swaps, transformations, text overlays, or watermarks.

\medskip
A realistic 60-second single continuous shot from a moving dashcam records a large urban intersection in daylight. Keep 13 road vehicles visible at detector-friendly scale, including five anchors: a red bus, a white SUV, a blue taxi, a silver cargo truck, and a black sedan. The five anchors remain continuously visible and retain the same colors, body types, proportions, and details. During 0-10 seconds, establish every anchor in normal traffic. During 10-25 seconds, the signal changes and traffic begins crossing; vehicles cross in depth with only brief partial overlap. During 25-40 seconds, turning cars cross the camera view while a bus continues straight, while the camera viewpoint changes smoothly and realistic spacing is maintained. During 40-52 seconds, the anchors complete another gradual merge or ordering change, briefly overlap, and reappear as the same vehicles. During 52-60 seconds, traffic continues naturally and all five anchors remain recognizable through the final frame. No collisions, cuts, teleportation, duplication, disappearance, vehicle transformation, or abrupt camera motion.
\end{consbenchprompt}

\begin{consbenchprompt}{Helios \textbar\ 60\,s  \textbar\ Example 2}
\label{prompt:helios-ex2}
Generate one realistic 60-second video in a single continuous, uninterrupted shot at normal speed. Keep 10-15 subjects visible at detector-friendly scale in a wide or medium-wide view. At least five distinctive anchor subjects must remain continuously visible and preserve identity, color, shape, scale, and markings for the full minute. Use one smooth camera trajectory and physically plausible motion. No cuts, transitions, camera switches, replay, montage, time jumps, slow motion, abrupt viewpoint changes, severe occlusion, tiny distant subjects, teleportation, duplication, disappearance, identity swaps, transformations, text overlays, or watermarks.

\medskip
A 60-second handheld camera video inside an open-air public plaza with fountains and seating areas. The scene contains 12 visible pedestrians. Five anchor pedestrians are a man wearing a white jacket with a black backpack, a woman wearing a red scarf and brown coat, a young person with curly hair wearing headphones, a man wearing a gray hat, and a woman wearing a blue dress carrying a small bag. During 0-10 seconds, pedestrians walk around the plaza, some stop briefly near benches while others continue moving. During 10-25 seconds, people cross in front of each other, two pedestrians temporarily overlap, and the camera operator walks around a fountain causing viewpoint changes. During 25-40 seconds, anchor pedestrians remain visible and maintain consistent facial appearance, clothing colors, and accessories. Single continuous wide-angle shot. During 40-52 seconds, the five anchor pedestrians take part in a second natural crossing and grouping interaction, briefly overlap, then separate with the same clothing, accessories, faces, and body proportions. During 52-60 seconds, normal walking continues and all five anchor pedestrians remain visible, recognizable, and unchanged through the final frame.
\end{consbenchprompt}

\section{Curriculum Training}
\label{app:curriculum}

\subsection{Supervised Fine-Tuning of 5s Models}
\label{app:5s-finetune}
We fine-tune the pretrained Wan-2.1-1.3B and SANA-2B using the same dataset and training protocol. The dataset comprises 72,000 nominal five-second OpenDV clips, totaling approximately 100 hours. A fixed split with seed 314159 partitions the data at the parquet-file level into 68,400 training clips and 3,600 validation clips (95\%/5\%). Each input contains 77 frames at a resolution of $480\times832$ (height $\times$ width), corresponding to 20 latent temporal positions. We use precomputed video latents and update all transformer parameters with full bidirectional attention.

Given a clean latent $\boldsymbol{z}$, Gaussian noise $\boldsymbol{\epsilon}$, noise level $\sigma$, and text condition $\boldsymbol{c}$, we optimize the conditional flow-matching objective:
\begin{equation}
\boldsymbol{z}_\sigma=(1-\sigma)\boldsymbol{z}+\sigma\boldsymbol{\epsilon},\qquad
\mathcal{L}=\mathbb{E}\left[
\frac{1}{D}\left|\boldsymbol{v_\theta}(\boldsymbol{z}_\sigma,\sigma,\boldsymbol{c})
-(\boldsymbol{\epsilon}-\boldsymbol{z})\right|_2^2\right],
\end{equation}
where $D$ denotes the number of latent elements per sample. We sample scheduler indices uniformly, use a flow shift of 3, and disable conditioning dropout.

\begin{table}[h]
  \centering
  \small
  \renewcommand{\arraystretch}{1.5}
  \setlength{\tabcolsep}{6pt}
  \caption{Settings for 5s, 10s, and 30s fine-tuning.}
  \label{tab:opendv-finetuning-settings}
  \resizebox{0.7\linewidth}{!}{%
  \begin{tabular}{@{}l|ccc@{}}
    \noalign{\hrule height 0.8pt}
    \textbf{Setting}
      & \textbf{5s} & \textbf{10s} & \textbf{30s} \\
    \hline
    Training clips
      & 68,400 & 34,208 & 24,016 \\
    Validation clips
      & 3,600 & 1,792 & 48 \\
    \hline
    Optimizer
      & \multicolumn{3}{c}{AdamW} \\
    Adam betas
      & \multicolumn{3}{c}{$(0.9, 0.999)$} \\
    Adam epsilon
      & \multicolumn{3}{c}{$10^{-8}$} \\
    Learning-rate schedule
      & \multicolumn{3}{c}{Cosine decay with linear warm-up} \\
    Weight decay
      & \multicolumn{3}{c}{0.01} \\
    Gradient clipping
      & \multicolumn{3}{c}{1.0} \\
    Training epochs
      & \multicolumn{3}{c}{1} \\
    \hline
    Peak learning rate
      & $10^{-5}$ & $3\times10^{-6}$ & $3\times10^{-6}$ \\
    Warm-up steps
      & 428 (5\%) & 107 (5\%) & 150 (10\%) \\
    Training updates
      & 8,550 & 2,138 & 1,501 \\
    Clips per validation
      & 128 & 128 & 32 \\
    \noalign{\hrule height 0.8pt}
  \end{tabular}%
  }
\end{table}

\begin{figure*}[h]
  \centering
  \includegraphics[width=0.85\textwidth]{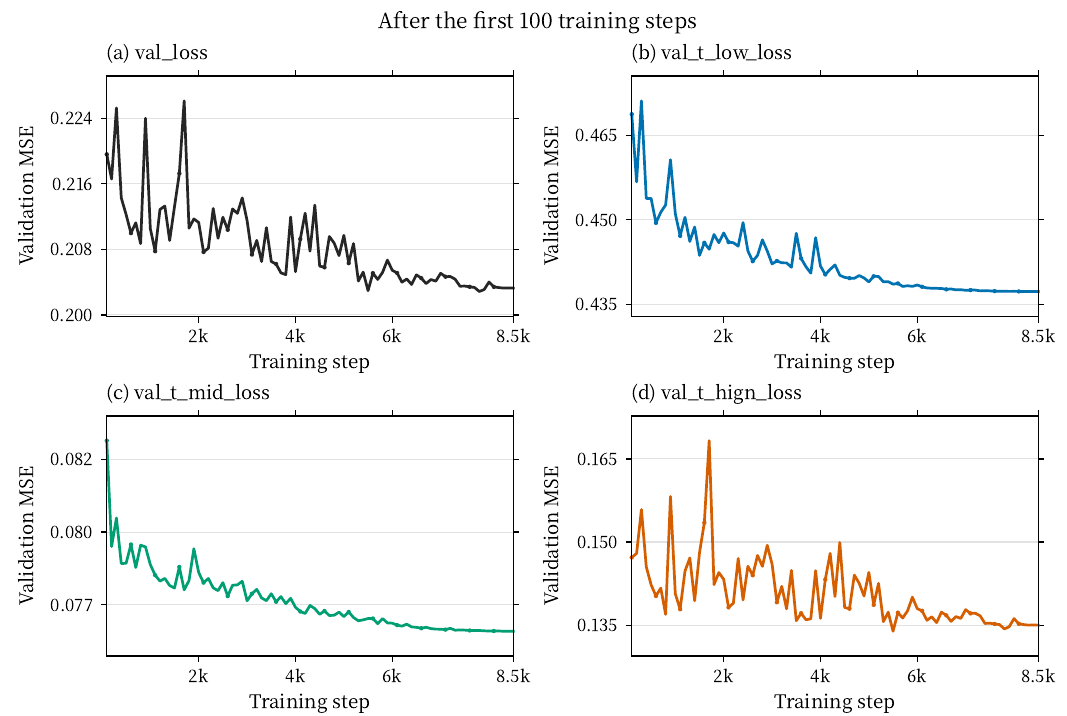}
  \caption{\textbf{Validation loss during 5s Wan-2.1-1.3B fine-tuning.}}
  \label{fig:opendv-5s-val-loss}
\end{figure*}

\begin{figure*}[h]
  \centering
  \includegraphics[width=0.85\textwidth]{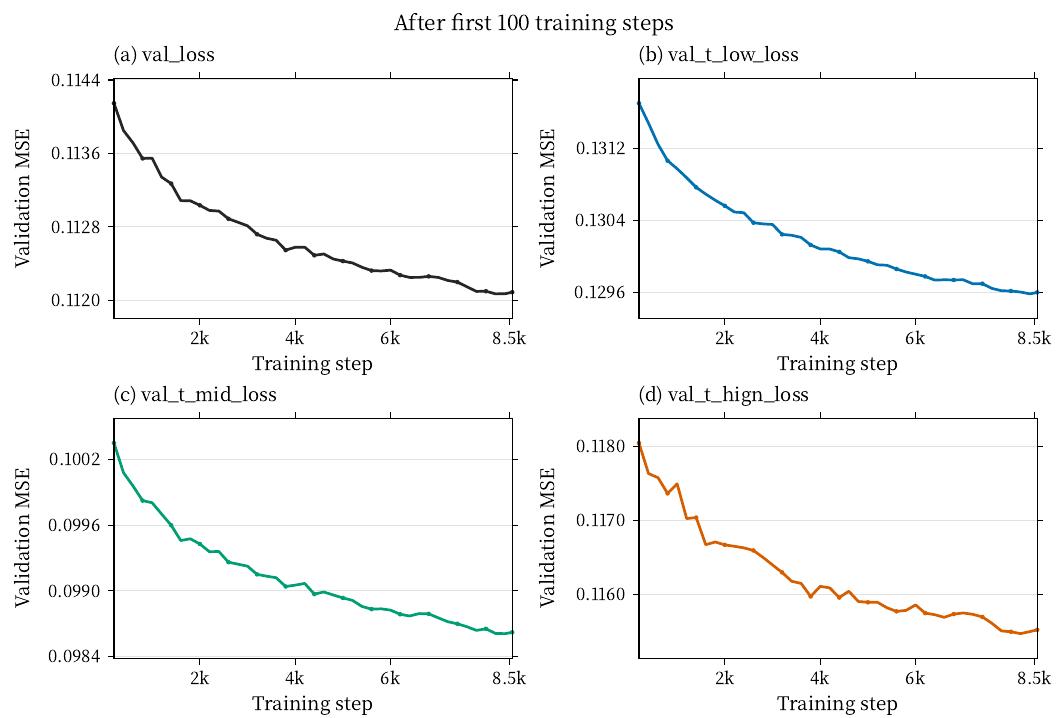}
  \caption{\textbf{Validation loss during 5s SANA-2B fine-tuning.}}
  \label{fig:sana-5s-val-loss}
\end{figure*}

Both models are trained for one epoch (8,550 updates) on eight A100-GPUs, with one sample per GPU and no gradient accumulation, yielding an effective batch size of eight. The learning rate increases linearly over the first 428 updates and subsequently follows a cosine decay. To monitor training stability, we evaluate 128 fixed held-out clips at initialization and every 100 updates, using fixed noise samples and timesteps with seed 314159. We record the overall velocity MSE and its low-, mid-, and high-noise components. Fig.~\ref{fig:opendv-5s-val-loss} and Fig.~\ref{fig:sana-5s-val-loss} show the training curves for Wan-2.1-1.3B and SANA-2B, respectively. Both models exhibit stable training and reduced validation loss. 
However, validation velocity MSE alone does not reliably indicate generated video quality: despite these improvements in both models, our final consistency evaluation and visual inspection favor the fine-tuned Wan-2.1-1.3B model. Given its stronger empirical performance and our limited computational budget, we use Wan-2.1-1.3B exclusively for subsequent fine-tuning experiments at 10s and 30s.

\subsection{Extending to Longer Videos}
\label{app:curriculum-long}

\begin{figure}[t]
  \centering
  \includegraphics[width=0.9\linewidth]{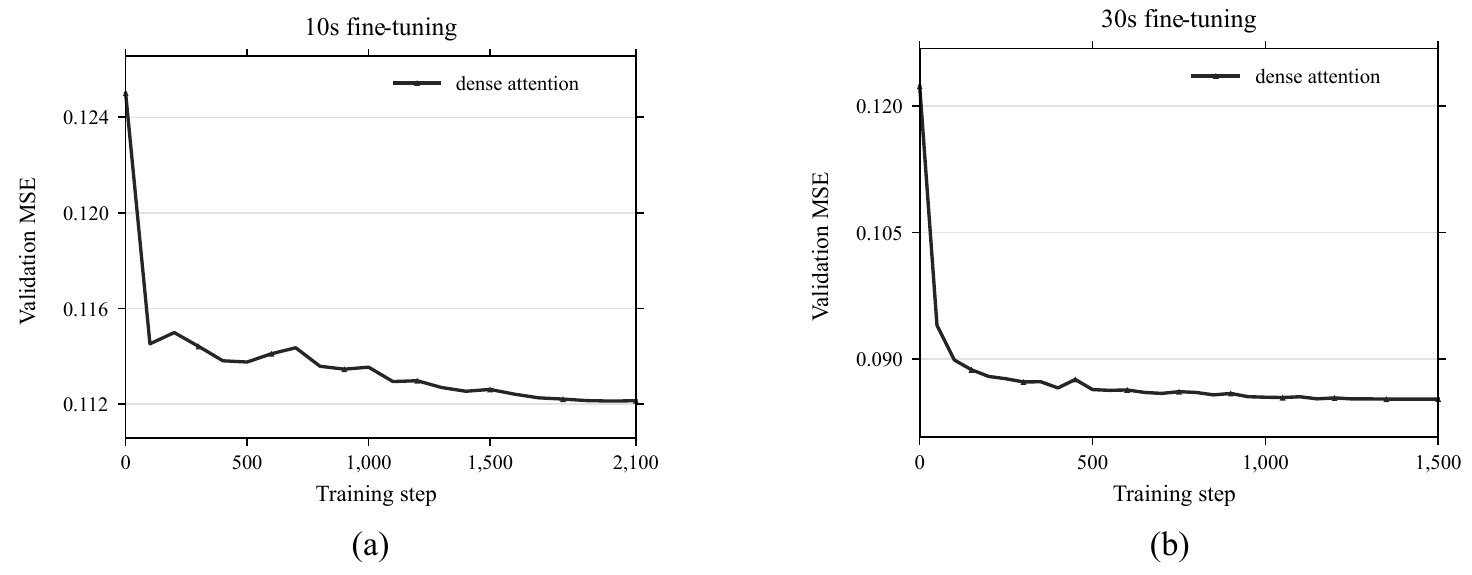}
  \caption{\textbf{(a). Validation loss during 10s fine-tuning. 
  (b). Validation loss during 30s fine-tuning.}}
  \label{fig:wan-long-val-loss}
\end{figure}

Fine-tuning on 10s and 30s clips follows the same optimization setup as the 5s stage, with the corresponding hyperparameters summarized in Tab.~\ref{tab:opendv-finetuning-settings}. We use dense spatiotemporal attention for both durations to enable information exchange across the entire video sequence. The 10s experiments are conducted on two nodes with eight A100 GPUs each, while the 30s experiments use four such nodes. As shown in Fig.~\ref{fig:wan-long-val-loss}, the validation loss decreases steadily for both durations and gradually stabilizes toward the end of training, indicating stable optimization when extending the model from 5s to longer video generation.

\section{Details of Consistency Post-Training}
\label{app:post-training}

\subsection{Implementation Details of Consistency Post-Training}
\label{app:post_training}

\paragraph{Policy parameterization.}
We freeze the base diffusion transformer and optimize only LoRA adapters inserted into the \texttt{to\_q}, \texttt{to\_k}, \texttt{to\_v}, and \texttt{to\_out} attention projections. The LoRA rank is $32$, with a scaling factor of $\alpha=64$. This lightweight setup aims to improve temporal consistency while preserving the model's overall generative capabilities and maintaining efficient training.

\paragraph{Three-policy optimization.}
We maintain three policies during training. The \emph{current policy} contains the trainable LoRA parameters and is optimized using AdamW. The \emph{old policy} generates training rollouts and is updated from the current policy using an exponential moving average after each optimizer step. The \emph{reference policy} is the frozen domain-adapted model obtained by disabling the LoRA adapters. The old policy stabilizes rollout generation, while the reference policy constrains the optimized model from deviating excessively from its original generation behavior. For each prompt, the old policy generates a group of $G=8$ videos. Their rewards are standardized within the group, enabling the current policy to learn from relative consistency differences under the same conditioning prompt. In the 5s setting, we accumulate 4 prompt groups before each optimizer update, resulting in 32 rollouts per update. Each rollout uses 10 diffusion sampling steps. For efficient optimization, we stratify the rollout trajectory and sample 4 diffusion timesteps for computing the training objective.

\begin{table*}[htbp]
\centering
\small
\caption{\textbf{Ablation on regularization for Wan-2.1-1.3B on 5s generation.}
Both optimized variants use the same combined consistency reward and are
evaluated at Step 75. Regularization consistently improves DynSC-Eval
metrics while preserving motion dynamics. DynSC-Eval metrics use four
significant figures, and VBench scores use a 0--100 scale.}
\label{tab:ablation_reg}
\resizebox{0.82\linewidth}{!}{%
\begin{tabular}{l*{10}{c}}
\toprule
& \multicolumn{6}{c}{\textit{DynSC-Eval Metrics}}
& \multicolumn{4}{c}{\multirow{2}{*}{\textit{VBench Metrics}}}
\\
\cmidrule(lr){2-7}
& \multicolumn{3}{c}{\textit{Local Consistency}}
& \multicolumn{3}{c}{\textit{Global Consistency}}
& \multicolumn{4}{c}{}
\\
\cmidrule(lr){2-4}
\cmidrule(lr){5-7}
\cmidrule(lr){8-11}
Setting
& SR$\downarrow$
& ER$\downarrow$
& MAD$\downarrow$
& ED$\downarrow$
& MRD$\downarrow$
& WFD$\downarrow$
& Mot$\uparrow$
& Dyn$\uparrow$
& Aes$\uparrow$
& Img$\uparrow$
\\
\midrule
\rowcolor{gray!10}
5s baseline
& 0.2560 & 9.177 & 0.1059
& 0.3837 & 0.2393 & 0.4428
& 98.35 & 92.00 & 50.47 & 71.40
\\
\midrule

w/o Reg.
& 0.2397 & 7.853 & 0.09514
& 0.3514 & 0.2223 & 0.4096
& 98.36 & 91.00 & 51.10 & 73.13
\\
\rowcolor{blue!10}
w/ Reg.
& \textbf{0.2220} & \textbf{7.139} & \textbf{0.08829}
& \textbf{0.3446} & \textbf{0.2129} & \textbf{0.4002}
& 98.49 & 92.00 & 50.81 & 71.82
\\

\bottomrule
\end{tabular}%
}
\end{table*}

\begin{figure}[tb]
    \centering
    \includegraphics[width=\textwidth]{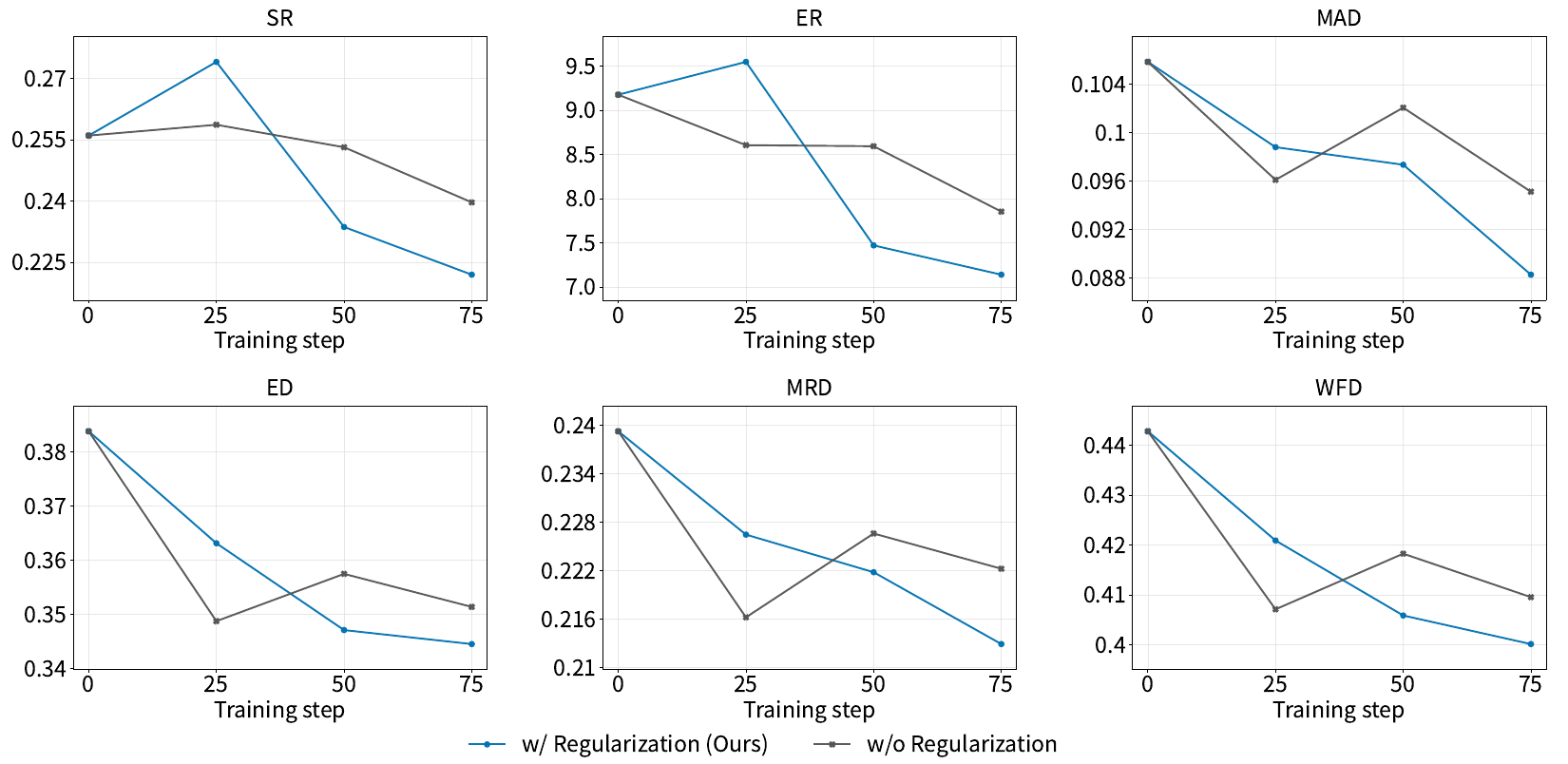}
    \caption{ \textbf{Effect of reference regularization on video consistency during fine-tuning}}
    \label{fig:rl-training-curve}
\end{figure}

\paragraph{Reference regularization.}
Optimizing consistency alone may encourage degenerate solutions, such as motion suppression or excessive deviation from the pretrained model. We therefore regularize the current policy against a frozen reference policy using spatial and temporal constraints. Table~\ref{tab:ablation_reg} compares consistency optimization with and without regularization and Fig.~\ref{fig:rl-training-curve} shows the validation curves of all six metrics. Even without regularization, optimizing the DynSC-Eval reward improves all six consistency metrics over the 5s baseline. Adding regularization further improves both local and global metrics, decreasing SR from $0.2397$ to $0.2220$ and ER from $7.853$ to $7.139$, while also improving MAD, ED, MRD, and WFD. Importantly, these gains do not come from suppressing motion: the VBench motion score remains nearly unchanged ($98.36$ vs.\ $98.49$), while dynamic degree increases from $91.00$ to $92.00$.  This indicates that regularization stabilizes reward optimization and yields additional consistency improvements without sacrificing video dynamics. 

\end{document}

%% file: math_commands.tex
\usepackage{amsmath,amsfonts,bm}

\def\eqref#1{equation~\ref{#1}}

\def\1{\bm{1}}

\DeclareMathAlphabet{\mathsfit}{\encodingdefault}{\sfdefault}{m}{sl}
\SetMathAlphabet{\mathsfit}{bold}{\encodingdefault}{\sfdefault}{bx}{n}

\def\sT{{\mathbb{T}}}

\def\sV{{\mathbb{V}}}

